\documentclass[letterpaper]{article} % DO NOT CHANGE THIS
\usepackage{aaai2026}  % DO NOT CHANGE THIS
\usepackage{times}  % DO NOT CHANGE THIS
\usepackage{helvet}  % DO NOT CHANGE THIS
\usepackage{courier}  % DO NOT CHANGE THIS
\usepackage[hyphens]{url}  % DO NOT CHANGE THIS
\usepackage{graphicx} % DO NOT CHANGE THIS
\usepackage{natbib}  % DO NOT CHANGE THIS AND DO NOT ADD ANY OPTIONS TO IT
\usepackage{caption} % DO NOT CHANGE THIS AND DO NOT ADD ANY OPTIONS TO IT
\usepackage{algorithm}

\usepackage{newfloat}
\usepackage{listings}
\DeclareCaptionStyle{ruled}{labelfont=normalfont,labelsep=colon,strut=off} % DO NOT CHANGE THIS
\floatstyle{ruled}
\newfloat{listing}{tb}{lst}{}
\floatname{listing}{Listing}

\usepackage{multirow}
\usepackage{tabularx}
\definecolor{softgreen}{rgb}{0.5, 0.8, 0.5}
\usepackage{tikz}
\usepackage{amsthm}
\usepackage{amssymb}
\usepackage{mathtools}
\usepackage{amsmath}
\newtheorem{definition}{Definition}
\newtheorem{zero hypothesis}{Null Hypothesis}
\usepackage{booktabs}
\usepackage{algpseudocode}
\usepackage{tcolorbox}
\newcommand\insertimptext[1]{
\begin{center}
\begin{tikzpicture}
    \node[draw=black, line width=1.5pt, rounded corners=5pt, fill=gray!10, inner sep=10pt, align=left, text width=0.9\columnwidth] 
 {\footnotesize\textbf{Takeaway:} \\ \parbox{\columnwidth}{#1 }};
\end{tikzpicture}
\end{center}
}
\title{Key Decision-Makers in Multi-Agent Debates: Who Holds the Power?}
\author{
    Qian Zhang\textsuperscript{\rm 1}, \ 
    Yan Zheng\textsuperscript{\rm 2} \thanks{Yan Zheng is the corresponding author.},\ 
    Jinyi Liu\textsuperscript{\rm 2},\ 
    Hebin Liang\textsuperscript{\rm 2},\ 
    Lanjun Wang\textsuperscript{\rm 1}
}
\affiliations{
    \textsuperscript{\rm 1}School of New Media and Communication, Tianjin University, China\\
    \textsuperscript{\rm 2}College of Intelligence and
Computing, Tianjin University, China\\
  \{zhangqian0827,\ yanzheng,\ jyliu,\ lianghebin\}@tju.edu.cn,\  wang.lanjun@outlook.com
   
}

\begin{document}

\maketitle

\begin{abstract}
Recent studies on LLM agent scaling have highlighted the potential of Multi-Agent Debate (MAD) to enhance reasoning abilities. However, the critical aspect of role allocation strategies remains underexplored. In this study, we demonstrate that allocating roles with differing viewpoints to specific positions significantly impacts MAD's performance in reasoning tasks. Specifically, we find a novel role allocation strategy, ``Truth Last'', which can improve MAD performance by up to 22\% in reasoning tasks. To address the issue of unknown truth in practical applications, we propose the Multi-Agent Debate Consistency (MADC) strategy,  which systematically simulates and optimizes its core mechanisms. MADC incorporates path consistency to assess agreement among independent roles, simulating the role with the highest consistency score as the truth. We validated MADC across a range of LLMs (9 models), including the DeepSeek-R1 Distilled Models, on challenging reasoning tasks. MADC consistently demonstrated advanced performance, effectively overcoming MAD's performance bottlenecks and providing a crucial pathway for further improvements in LLM agent scaling.
\end{abstract}

\begin{links}
    \link{Code}{https://github.com/SG-XM/AAAI2026-MADC}
    % \link{Extended version}{https://aaai.org/example/extended-version}
\end{links}

\section{Introduction}
\begin{figure}[!t]
  \includegraphics[width=1\columnwidth]{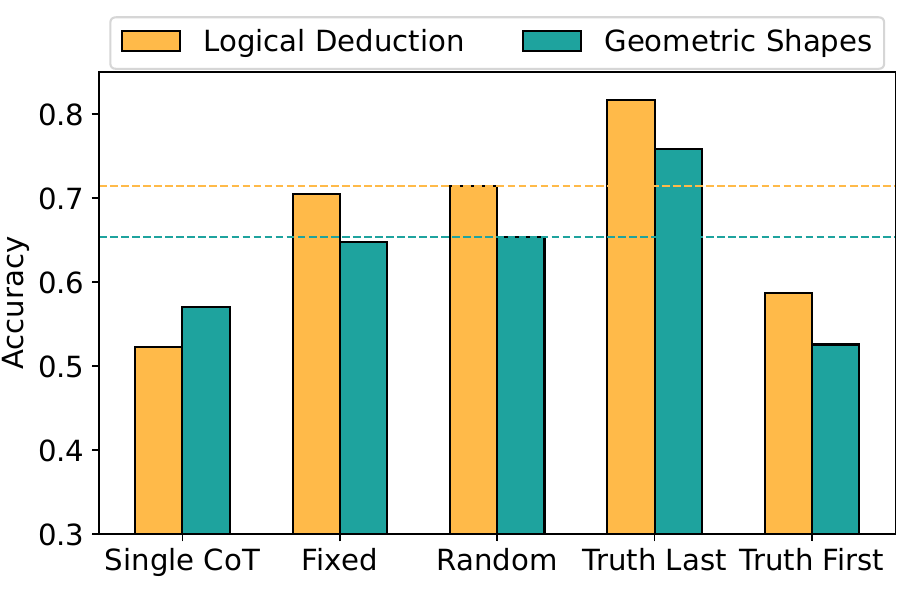}
  \caption{\small Comparative performance of single-agent and different MAD allocation strategies using the Qwen2.5-7B-Instruct model on BBH's Logical Deduction and Geometric Shapes tasks.}
  \label{fig:intro_1}
\end{figure}

Although Large Language Model (LLM) agents have shown remarkable performance in logical reasoning~\citep{shinn2023reflexion,deepseekai2024deepseekv3technicalreport}, natural language processing tasks and  specific vertical domains~\citep{liu2025chaosorderatomicreasoner,xiao2024cellagent,chen2024sheetagent,reddy2024infogent}, they struggle with more complex tasks~\citep{gou2024critic,chen2024agentverse}. To further improve the performance, numerous multi-agent frameworks have been proposed, such as CAMEL~\citep{li2023camel},  LLM-Blender~\citep{jiang-etal-2023-llm}, CMD~\citep{wang-etal-2024-rethinking-bounds}, Debate~\citep{liang-etal-2024-encouraging}, CIPHER~\citep{pham2024let}, Socialized Learning~\citep{yao2024socialized}, RECONCILE~\citep{chen-etal-2024-reconcile}. Especially, some studies have discovered the scaling laws of agents~\citep{kaplan2020scaling,wang2025mixtureofagents}.

\begin{figure*}[!h]
  \includegraphics[width=1\linewidth]{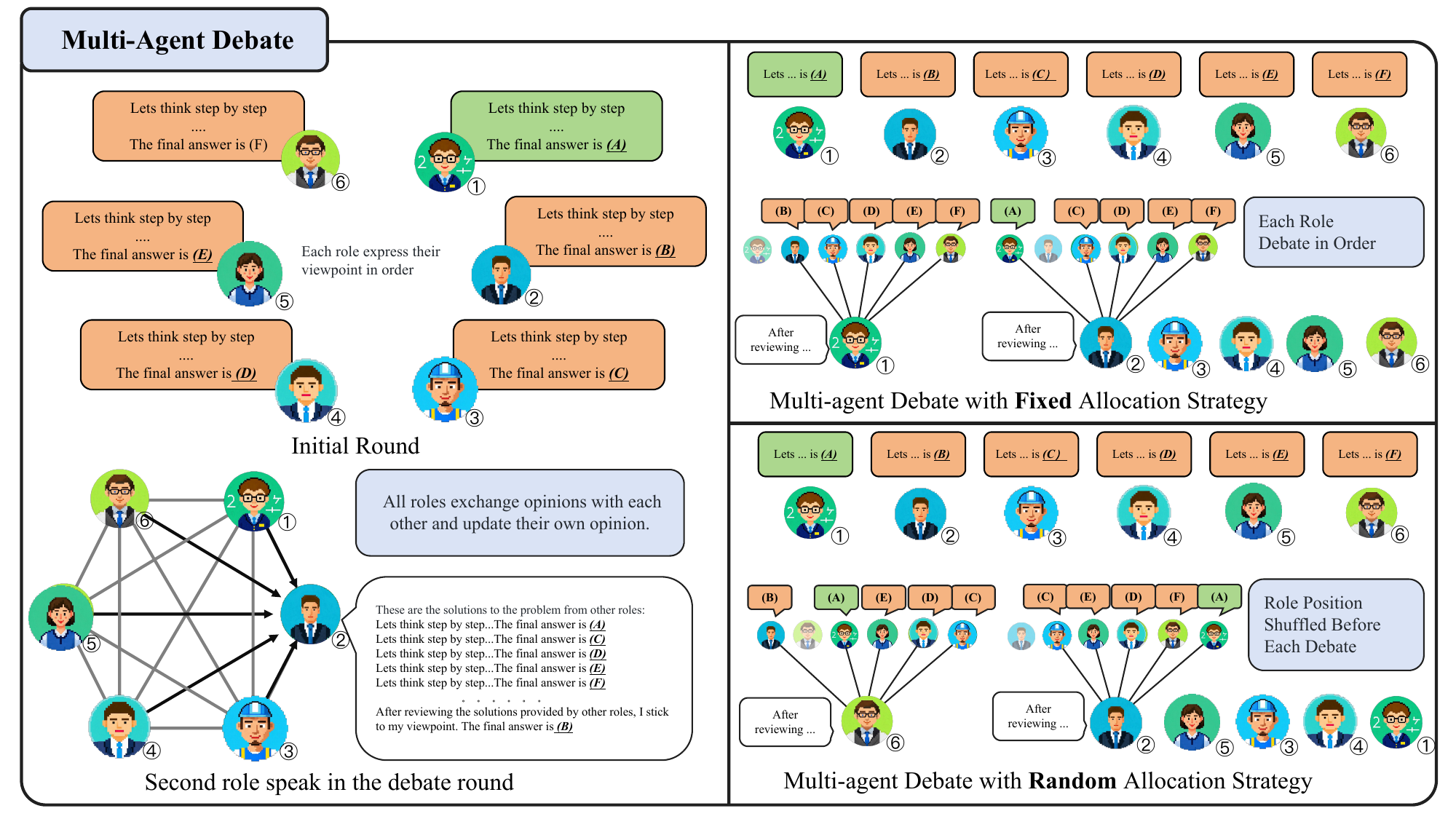} 
  
  \caption {In the initial round of the MAD framework, each role independently uses CoT. During the debate round, roles exchange viewpoints in a fully connected manner to update their viewpoints. In the default Fixed strategy, roles speak in a consistent order each round and receive others' viewpoints in a fixed sequence (A)-(E). The Random strategy disrupts both the speaking order and the relative positions during the debate. Green represents correct viewpoints, while orange indicates incorrect viewpoints.}
  \label{fig:pre_1}
\end{figure*}

Recent studies explore multi-agent scaling principles. Du \citep{du2023improving} proposed a multi-agent debate framework, showing performance gains with more agents and debate rounds. Agent Forest \citep{li2024more} scales agents to boost LLM results, aligned with task difficulty. And traditionally, some approaches improve performance through post-trian \citep{cheng-etal-2025-dualrag}. And some approaches in MAS have leveraged reinforcement learning (RL) to enhance performance \citep{DBLP:journals/corr/abs-2109-06668,yang2025dual,8952543}.

However, stable scaling faces challenges: Wang \citep{wang-etal-2024-rethinking-bounds} found strong single-agent prompts match top multi-agent methods on many reasoning tasks; Smit \citep{smitshould} noted multi-agent debating systems don’t consistently outperform strategies like self-consistency (though some MAD systems improve with hyperparameter tuning). Li \citep{li-etal-2024-improving-multi} showed more reference solutions aid easier questions but mislead on harder ones, reducing accuracy. These reveal MA scaling instability due to parameter sensitivity and task complexity.

Figure \ref{fig:intro_1}  presents a comparison of the performance of Single-Agent and Multi-Agent Debate frameworks with varying role allocation strategies. In this figure, `Single' represents the performance of a single agent using Chain of Thought (CoT)~\citep{wei2022chain}, while `MAD' refers to the framework proposed by Du~\citep{du2023improving}. In this context, `Fixed' indicates roles placed in predetermined positions, `Random' denotes roles arranged in a random order during the debate, `Truth Last' refers to arranging agents with correct viewpoints at the end, while `Truth First' does the opposite (Defined in \ref{sec:allocation_strategy}). Our observations reveal that with the Qwen2.5-7B-Instruct model, the existing fixed role allocation strategy outperforms the single agent. And there are significant differences between various role allocation strategies.

Based on observations from Figure \ref{fig:intro_1}, we conducted systematic experiments for in-depth analysis. Although many MA discussion, debate, and collaboration frameworks exist, they often overlook role allocation strategies. For example, in Du's MAD framework, role positions are fixed in the implementation shown in Figure \ref{fig:pre_1}.

We found that when role allocation are more ordered by viewpoints, debate results show lower entropy, without impacting the accuracy of conclusions. Our results also demonstrate that using a white-box allocation strategy "Truth Last" significantly enhances accuracy in reasoning tasks, and this improvement is orthogonal to existing MAD frameworks, requiring no changes to context or prompts. Then we propose a Multi-Agent Debate Consistence allocation strategy to simulate the ``Truth Last". This method effectively enhances MAD performance and efficiency. Additionally, we discovered that under ideal conditions, only half the number of agents is needed to reverse a debate outcome, further deepening our understanding of role allocation strategies in scaling laws.

In summary, our contributions are:

(1) Key Insight: We identify role allocation strategy as a novel scaling dimension for multi-agent debate frameworks, revealing that later-positioned roles exert greater influence on final decisions through analysis of debate entropy and positional bias.

(2) Methodological Innovation: We propose the Multi-Agent Debate Consistency (MADC) allocation strategy, a novel approach that optimizes role arrangement without modifying prompts or context, effectively unlocking MAD's potential. This method is completely \textbf{orthogonal} to existing MA frameworks.

(3) Empirical Validation: We evaluated 9 different LLM models, including open-source models like Qwen2.5, closed-source models like GPT-4o-mini, and new architectures like DeepSeek-R1-Distill. Our comprehensive tests across three diverse tasks, logical deduction, geometric shapes, and MATH500, which highlight the broad applicability and robustness of our findings.
\section{Preliminary} \label{sec:pre}
To study the impact of allocation strategy on the multi-agent (MA) framework, we need to control other influencing factors. We must select the most representative parameter settings to ensure our findings are comprehensive and applicable.

\subsection{The Factors that Impacts the Performance of MAD}

In MAD, the factors affecting performance mainly include the following aspects.

\textbf{Communication Strategies:}
Existing work explores various strategies(example can be found in Appendix) such as Talk One-by-One, Simultaneous-Talk, Simultaneous-Talk-with-Summarizer~\cite{chan2024chateval}, exchanging natural language, and exchanging raw transformer output embeddings~\cite{pham2024let}. In our study, we choose the basic Simultaneous-Talk strategy and exchange natural language.

\textbf{Topology and Connectivity:}
Another key factor is the topology and connectivity among agents. Some studies propose group discussion structures~\cite{wang-etal-2024-rethinking-bounds}, tree structures, and dynamically adjusting the topology by removing agents during discussions~\cite{liu2024a}. In our work, we choose the traditional method of full connectivity, where all agents communicate with each other.

\textbf{Agents Profiling:}
Profiling prompts are also a critical factor influencing multi-agent (MA) systems~\cite{hong2023metaGPT,li2023you,chen2023autoagents,park2023generative,park2024generative}. Common prompts settings include collaborative~\cite{zhang-etal-2024-exploring,10.24963/ijcai.2024/867}, adversarial~\cite{chan2024chateval} and judgment prompt~\cite{wang-etal-2024-rethinking-bounds}. In our work, we choose not to use special agents profiling to eliminate their influence.

\textbf{Role Allocation Strategies:}\label{sec:allocation_strategy}
% Our goal is to falsify the hypothesis and compare debate outcomes across different positions to identify the optimal positioning.
Based on the parameter settings mentioned above, we will experiment within the Multi-Agent Debtate (MAD) framework~\cite{du2023improving} .
Based on MAD, we have defined various role allocation strategies during the debate process. \textbf{(1) Fixed}: The position of roles is fixed once allocated. \textbf{(2) Random}: Roles are randomly allocated before each debate.\textbf{(3) Truth First}: Roles with the correct answer are allocated at the beginning. \textbf{(4) Truth Last}: Roles with the correct answer are allocated at the end.

\subsection{The Formulation about MAD Scaling}
In the MAD framework, we define \( n \) agents \( A_1, A_2, \ldots, A_n \), each generating a viewpoint \( V_{i,j} \) in every round of debate, where \( i \) denotes the agent index and \( j \) represents the debate round. 

Figure \ref{fig:pre_1} illustrates MAD framework. During each round of discussion and debate, the agents maintain fixed positions throughout. For instance, in the first round, $A_0$ to $A_5$ express $V$ as follows: (A), (A), (A), (B), (C), (D). In subsequent rounds, the agent with viewpoint D consistently follows the agent with viewpoint C during discussions and debates. In a multi-persona context, this means that the agent corresponding to $role_5$ always follows $role_4$. Practically, this could result in a critic always speaking after an advisor, or a programmer consistently following a product manager.

\section{Methodology}
\subsection{Does Role Allocation Strategy Impact MAD?}
The conclusions from Figure \ref{fig:intro_1} are clear but not comprehensive or robust enough. Firstly, we propose a null hypothesis, which we will attempt to falsify:
\begin{zero hypothesis}[Non-impactfulness, $H_0$]
The position of roles has no impact on the final outcome in a debate.
\end{zero hypothesis}

To falsify \( H_0 \), we must demonstrate that there exists at least one allocation strategy, \( S \), such that the outcome using \( S \) differs significantly from outcomes with Fixed or Random strategy. 
\subsection{Challenging the Null Hypothesis: Impactful Allocation Strategies}

To find sufficient evidence to reject $H_0$, we evaluated the accuracy metric of MAD after one round of debate under different allocation strategies. Additionally, we examined the log-likelihood(Appendix A.1) and entropy (Appendix A.2) metrics. Accuracy represents the number of correctly answered questions, while log-likelihood indicates the number of roles with correct answers.

\begin{figure}[!tb]
\centering
  \includegraphics[width=\columnwidth]{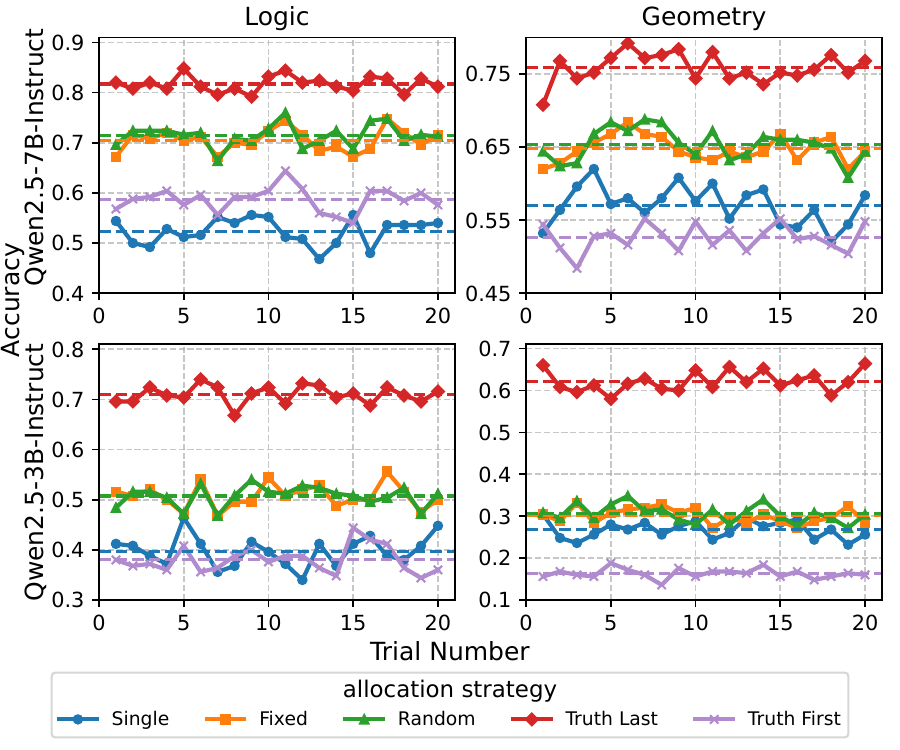}
  \caption{Experimental results showing the accuracy metrics across different allocation strategies, with experiments repeated 20 times.}
  \label{fig:null_1}
\end{figure}

To test this, we conducted 20 repeated experiments using four
allocation strategies outlined in Section \ref{sec:pre}, focusing on Logical Deduction and Geometric Shapes tasks. The results are shown in Figure \ref{fig:null_1}.

We observed no significant difference between the random and fixed strategies in the experiments, which is reasonable since the fixed strategy can be seen as a single sample of the random strategy. Noting that the Truth First strategy outperformed others significantly with p-value $\textless$0.0001. 
\begin{figure}[!htb]
  \includegraphics[width=\columnwidth]{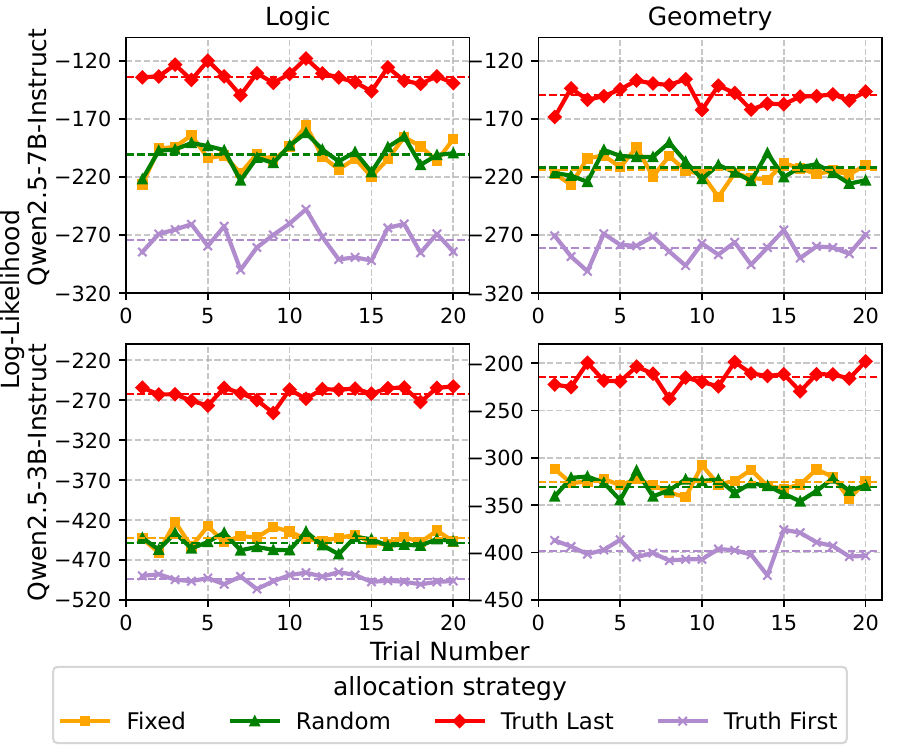}
  \caption{Experimental results showing the log-likelihood metrics across different allocation strategies, with experiments repeated 20 times.}
  \label{fig:null_2}
\end{figure}

Figures \ref{fig:null_1} and Figure \ref{fig:null_2} demonstrate that the Truth Last strategy significantly outperforms other strategies, which disproves the $H_0$. 
To further challenge the $H_0$, we examine how allocation strategies affect viewpoint convergence. Lower entropy reflects higher consensus (which is explained in Appendix A.2), suggesting that role allocation strategy affects debate process and thus challenges the $H_0$.

\begin{figure}[!htp]
  \includegraphics[width=\columnwidth]{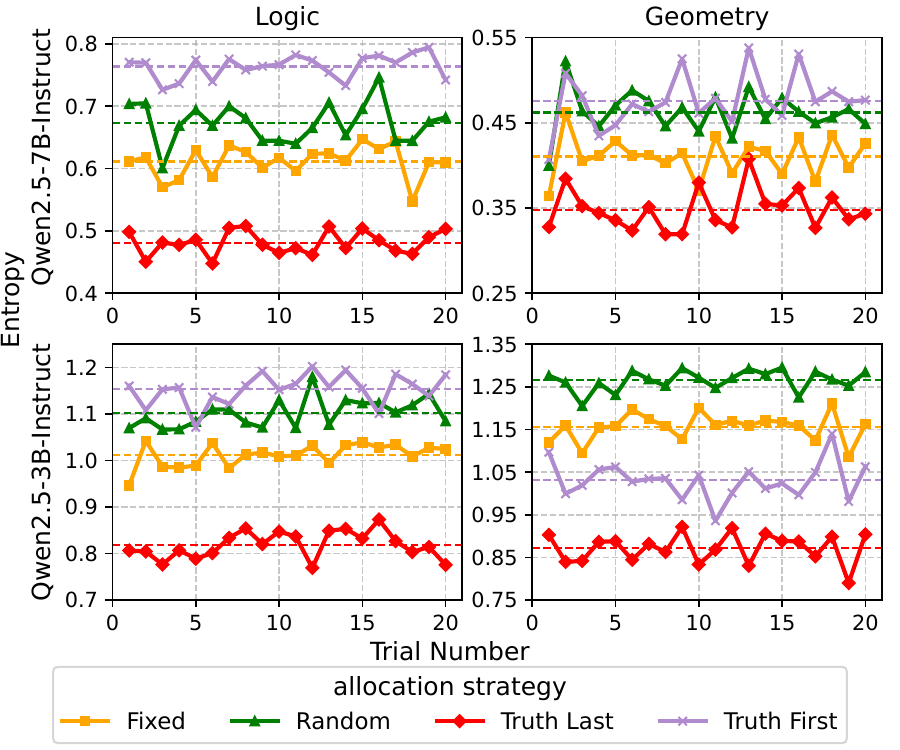}
  \caption{Experimental results showing the entropy metrics across different allocation strategies, with experiments repeated 20 times.}
  \label{fig:null_3}
\end{figure}

Figure \ref{fig:null_3} presents the entropy metrics for various strategies, showing interesting results: the Truth Last strategy has the lowest entropy, followed by fixed, and then random. Truth First, although resulting in lower accuracy, makes it easier for debates to reach consensus, showing lower entropy. For example, on the geometry dataset, the 3B model with Truth First exhibited lower entropy compared to random and fixed strategies. Despite similar accuracy and log-likelihood metrics, random and fixed strategies show significant differences in entropy, further challenging the null hypothesis. Additionally, the Truth Last can be seen as a sorted version of fixed, demonstrating that entropy decreases as the orderliness of agent allocation increases.

\insertimptext{

(1) Allocating the role with the correct viewpoint last can enhance the effectiveness of MAD.

(2) An orderly allocation of roles is more conducive to reach an agreement in the debate.}

\subsection{Multi-Agent Debate Consistency Strategy}
In previous experiments, we demonstrated the importance of allocation strategies in MAD settings and found that Truth Last consistently yields better results. However, in practical applications, the truth is unknown. To address this, we propose a method called MADC to approximate the truth using confidence levels.

\paragraph{Path Consistency.}

After each round, agents may arrive at an intermediate result \( V_{i,j} \), with each result serving as a candidate for final decision-making.

\begin{definition}[Debating Path]
A Path \( P_i \) is the sequence of viewpoints or intermediate results generated by an agent \( A_i \) during the debate process. Formally, it is represented as:
\begin{equation}
    P_i = (V_{i,1}, V_{i,2}, \ldots, V_{i,m}),
\end{equation}
where \( V_{i,j} \) is the viewpoint of agent \( A_i \) at round \( j \), and \( m \) is the total number of rounds.
\end{definition}

For \( V_{i,j} \), when \( j < m \), \( V_{i,j} \) is considered an intermediate result, and when \( j = m \), it is the final result. For any \( V_{i,j} \), there exists a corresponding intermediate path \( P_{i,j} = (V_{i,1}, V_{i,2}, \ldots, V_{i,j}) \), representing the sequence of results up to that round.

\begin{definition}[Path Consistency]
Path Consistency measures how likely different paths produce the same result at a specific round j. Consider paths \( P_{1,j}, P_{2,j}, \ldots, P_{n,j} \). These are followed by agents \( A_1, A_2, \ldots, A_n \). Path Consistency is the probability that their results \( V_{1,j}, V_{2,j}, \ldots, V_{n,j} \) are equal. This occurs at round \( j \), where \( j < m \). We define it as:
\begin{equation}
\label{eq:consistency}
    \begin{aligned}
        &\text{Consistency}(\mathcal{P}_{i,j})= \sum_{k\neq i}\Pr(V_{i,j} = V_{k,j}).
    \end{aligned}
\end{equation}
Here, \( V_{i,j} \) is the result of agent \( A_i \) at round \( j \).
\end{definition}
To approximate the ground truth, we apply the MaxProb principle, which utilizes path consistency to enhance result accuracy. For each reasoning path \( P_{i,j} \), we calculate the consistency of its result \( V_{i,j} \) at round \( j \) as \( \text{Consistency}(P_{i,j}) \). The result corresponding to the path with the highest path consistency score is selected.:
\begin{equation}
V^* = \arg\max_{V_{i,j}} \text{Consistency}(P_{i,j}).
\end{equation}

\begin{algorithm}[t]
\caption{Multi-Agent Debate Consistency}
\label{alg:madc}
\begin{algorithmic}[1]
\Require Agent set $\mathcal{A} = \{A_1,...,A_n\}$ with initial paths $\{P_{i,1}\}_{i=1}^n$, total rounds $m$
\Ensure Consistency ordered list $\mathcal{L}$

\State Initialize path matrix $\mathbf{P}^1 \gets [P_{1,1} \cdots P_{n,1}]^\top$ \Comment{$P_{i,j}$: path of $A_i$ at round $j$}
\State Initialize consistency list $\mathcal{L}^1 \gets \emptyset$

\For{round $j \gets 1$ \textbf{to} $m-1$} \Comment{Consistency evaluation phase}
    \State Compute path consistency: 
    \State $\quad \Phi(P_{i,j}) \gets \Pr(V_{1,j} = \cdots = V_{n,j} | P_{i,j})$ \Comment{According to Eq. \ref{eq:consistency}}
    \State Identify max-consistency agent: 
    \State $\quad A^* \gets \arg\max_{A_i \in \mathcal{A}} \Phi(P_{i,j})$ 
    
    \State Update debate ordering:
    \State $\quad \mathcal{L}_{\text{temp}} \gets \text{argsort}(\{\Phi(P_{k,j}) | A_k \in \mathcal{A}\setminus A^*\})$ 
    \State $\quad \mathcal{L}^{j} \gets (\mathcal{L}_{temp} \circ A^*)$

    \State Propagate consistency: 
    \State $\quad \mathbf{P}^{j+1} \gets \text{Debate}(\mathbf{P}^j, \mathcal{L}^{j})$ 
\EndFor

\end{algorithmic}
\end{algorithm}

Path Consistency indicates the level of agreement among different agents' debating paths at a specific round. When the consistency is high, it suggests that multiple independent agents have converged on the same result. This convergence implies robustness and reliability, as it reduces the likelihood of individual biases or errors impacting the outcome.

\paragraph{Debating Strategy Guided by Path Consistency.}

Multi-Agent Debate Consistency (MADC) is designed to enhance the adaptability and effectiveness of multi-agent systems by leveraging consistency metrics during debates. One of the key advantages of MADC is its \textbf{orthogonality} to other multi-agent methods, meaning it can be integrated without altering existing prompts or context.

Algorithm~\ref{alg:madc} outlines the MADC process in a structured manner. Initially, consistency scores are calculated for each round using Equation~\ref{eq:consistency} (Lines 5-6). Agents are then reordered based on these scores (Lines 8-9). This reordering allows all agents to update their paths effectively (Line 11).

MADC integrates seamlessly with existing systems through two key features.
Its plug-in design introduces dynamic allocation without altering the internal mechanisms of the agents. Additionally, MADC promotes self-improvement by ensuring that more reliable viewpoints have a greater influence, thereby enhancing decision-making without modifying the existing framework.

\section{Experiments}

To validate the effectiveness of our method and gain deeper insights into how allocation strategies impact the MAD debate process and outcomes, we focus on the following research questions: 

\textbf{\textit{RQ1 (Applicability and Potential):}}Is MADC robust enough across various models and tasks?

\textbf{\textit{RQ2 (Scaling Impact):}}Is the allocation strategy still crucial in round and agent count scaling?

\textbf{\textit{RQ3 (Quantitative Analysis):}}Can just half the agents reverse a debate outcome?

\subsection{Experiment Setup}\label{sub:exp}

\paragraph{Language Model.}
To validate the MADC strategy, we tested multiple models\cite{OpenAI,Qwen,Qwen2.5,GLM}: GPT, Qwen GLM, DeepSeek. These models include both open-source and closed-source options. For closed-source models, we conducted experiments using API calls, covering both free and paid versions. And we adopted very simple prompt settings(refer to Appendix A.4).

\paragraph{Tasks and Dataset.} 

MATH500\cite{MATH-500}: A specialized dataset of 500 diverse mathematical problems spanning algebra, geometry, and number theory, requiring multi-step symbolic reasoning.

Logical Deduction: A subtask of Big-Bench Hard (BBH)\cite{suzgun2022challenging} focusing on deductive reasoning under constrained premises. It evaluates a model’s ability to infer implicit logical relationships.

Geometric Shapes: Another BBH subtask targeting spatial and geometric reasoning. Models must interpret text-based shape descriptions.

\begin{figure*}[!htb]
  \includegraphics[width=1\linewidth]{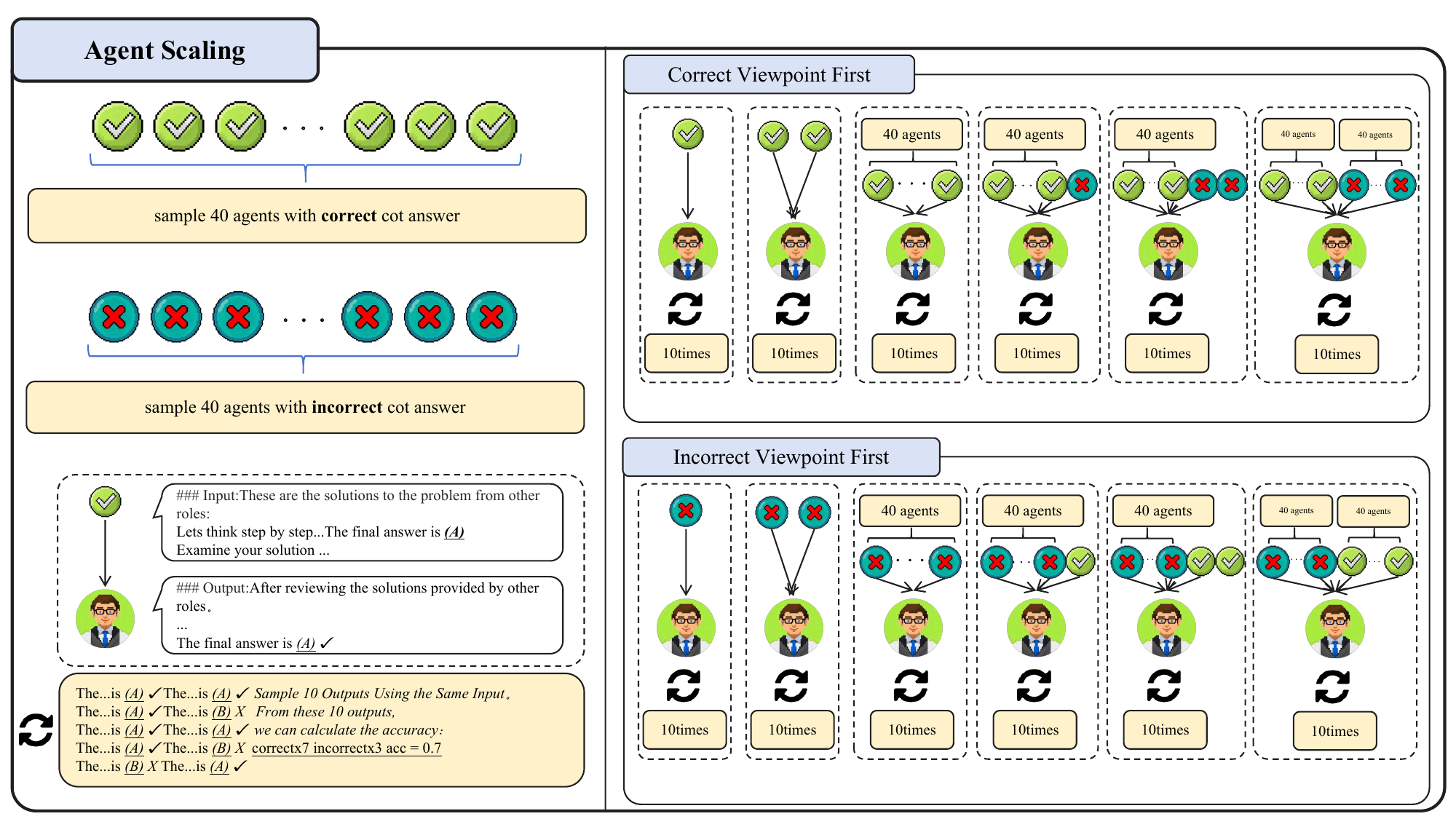} 
  
  \caption {For each question, we resample to obtain 40 agents with correct viewpoints and 40 with incorrect viewpoints. In the debate round, we fix the context and conduct 10 repetition for each number of agents, calculating the accuracy. We apply two scaling strategies. The entire experiment is repeated five times to obtain the average value in scaling.}
 \label{fig:case_2}
\end{figure*}
\paragraph{Settings(RQ3): Evaluating the Agent Threshold for Reversing Debate Outcomes.}

We constructed a specific scenario shown as Figure \ref{fig:case_2}. We constructed two groups of agents: 40 with correct viewpoints and 40 with incorrect viewpoints. In the most extreme scenario, we first allocated the 40 agents with correct viewpoints, then gradually added agents with incorrect viewpoints. Each configuration was tested in 10 debate repetitions to calculate accuracy under the current setup, and the entire experiment was repeated 5 times. For comparison, we implemented three scaling methods:

\textbf{Correct Viewpoint First}: Scaling agents with correct viewpoints first.

\textbf{Incorrect Viewpoint First}: Scaling agents with incorrect viewpoints first.

\textbf{Alternating Scale}: Alternately scaling agents with correct and incorrect viewpoints.

When constructing the agents, we used Qwen2.5-7B-Instruct to perform chain-of-thought (CoT) sampling 400 times for each question. This ensured a sufficient number of agents with diverse viewpoints.

\subsection{Performance of Role Allocation Strategy (RQ1)}
\begin{table*}[!hbt]
    \centering
    \caption{Performance of Role Allocation Strategy(10 agents 1 debate round)}
    \label{tab:madc_2}
    \scalebox{0.95}{
    \begin{tabular}{lllllllll}
        \toprule
        \textbf{Strategy}& \textbf{Model} & \textbf{Logic}    & \textbf{Geometry}  &\textbf{MATH500} & \textbf{Model}  & \textbf{Logic}    & \textbf{Geometry}  &\textbf{MATH500}  \\ 
        \midrule
         Single CoT        & \multirow{5}{*}{GPT-4o-mini} &70.0             & 63.2       &  69.8  &\multirow{5}{*}{\parbox{1.7cm}{Qwen2.5-3B-Instruct}} & 42.0              & 27.2             &61.4 \\
                                   Self-Consistency  &                & 82.4             & 66.4       &  76.2 && 54.0              & 28.8             &69.2      \\
                                    MAD   &              & 87.2             & 72.2       &  76.4& &41.2              & 25.2        &69.6 \\
                                    MAD+Truth & & 90.4 {\textbf{\scriptsize\textuparrow 3.2}}  & \textbf{76.0} {\textbf{\scriptsize\textuparrow 4.0}}   &76.4 \textbf{- -}& & \textbf{64.0} {\textbf{\scriptsize\textuparrow 22.8}}  & \textbf{48.0} {\textbf{\scriptsize\textuparrow 22.8}} &76.2 {\textbf{\scriptsize\textuparrow 6.6}}\\
                                    \textbf{MAD+MADC}  &   & \textbf{89.2} {\textbf{\scriptsize\textuparrow 2.0}}   & 73.2 {\textbf{\scriptsize\textuparrow 1.2}}   &76.4 \textbf{- -}& &\textbf{50.0} {\textbf{\scriptsize\textuparrow 8.8}}   & 28.4 {\textbf{\scriptsize\textuparrow 3.2}}   &71.0 {\textbf{\scriptsize\textuparrow 1.4}}\\ 

        \midrule
        Single CoT        &\multirow{5}{*}{GLM-4-Flash}         &  46.0              & 41.2         &43.2&\multirow{5}{*}{\parbox{1.7cm}{Qwen2.5-7B-Instruct}} & 52.4              & 57.0          &67.8\\
                                            SC  &                & 60.8              & 50.0         &54.6&& 60.2              & 64.0         &75.8\\
                                           MAD  &                & 58.4              & 50.8         &51.2&& 70.4              & 64.8         &74.4 \\
                                            MAD+Truth  &  & \textbf{72.8} {\textbf{\scriptsize\textuparrow 14.4}}  & 62.4 {\textbf{\scriptsize\textuparrow 11.6}} &53.2 {\textbf{\scriptsize\textuparrow 1.6}}&& \textbf{81.8} {\textbf{\scriptsize\textuparrow 11.4}}&   76.0 {\textbf{\scriptsize\textuparrow 11.2}}   &80.4 {\textbf{\scriptsize\textuparrow 6.0}}\\
                                            \textbf{MAD+MADC}  &   & \textbf{62.0} {\textbf{\scriptsize\textuparrow 3.6}}   & 52.4 {\textbf{\scriptsize\textuparrow 1.2}}    &52.2 {\textbf{\scriptsize\textuparrow 1.0}}&& 72.0 {\textbf{\scriptsize\textuparrow 1.6}}  & \textbf{68.4} {\textbf{\scriptsize\textuparrow 3.6}}   &76.8 {\textbf{\scriptsize\textuparrow 2.4}}\\ 
        \midrule
        Single CoT        &\multirow{5}{*}{GLM-4-FashX}        &  42.4             & 42.4        &42.8&\multirow{5}{*}{qwen-turbo}    & 71.2              & 53.2         &73.2\\
                                           SC  &                 & 57.2             & 50.8        &53.8&& 82.8              & 55.6         &76.8\\
                                           MAD  &                & 60.4             & 49.2        &53.4&& 82.4              & 58.4         &78.2\\
                                             MAD+Truth &  & 70.0 {\textbf{\scriptsize\textuparrow 9.6}}   & \textbf{62.0} {\textbf{\scriptsize\textuparrow 12.8}}  &55.2 {\textbf{\scriptsize\textuparrow 1.8}}&& \textbf{91.6} {\textbf{\scriptsize\textuparrow 9.2}}   & 64.8 {\textbf{\scriptsize\textuparrow 6.4}}  &80.2 {\textbf{\scriptsize\textuparrow 2.0}}\\
                                            \textbf{MAD+MADC}  &   & 61.6{\textbf{\scriptsize\textuparrow 1.2}}   & \textbf{51.2} {\textbf{\scriptsize\textuparrow 2.0}}  &55.0 {\textbf{\scriptsize\textuparrow 1.6}}&& 83.2 {\textbf{\scriptsize\textuparrow 0.8}}   & \textbf{60.4} {\textbf{\scriptsize\textuparrow 2.0}}  &78.2 \textbf{- -}\\ 
        \midrule
        Single CoT  &\multirow{5}{*}{\parbox{1.7cm}{DeepSeek-R1-Distill-Qwen-7B}}&  61.2     &62.4    &86.2  &\multirow{5}{*}{qwen-plus}  & 84.8              & 52.4         &77.0\\
                                             SC&                 &  86.8        & 74.0     &90.6&& 92.0              & 54.8         &80.0\\
                                             MAD  &              &  76.0        & 65.2     &87.2& & 94.0              & 65.2         &79.8\\
                                             MAD+Truth &  &  89.6 {\textbf{\scriptsize\textuparrow 13.6}}        & \textbf{89.6} {\textbf{\scriptsize\textuparrow 24.4}}   &91.8 {\textbf{\scriptsize\textuparrow 4.6}}& & 94.0 {\textbf{\scriptsize\textuparrow 0.0}}    & \textbf{69.6} {\textbf{\scriptsize\textuparrow 4.4}}  &80.8 {\textbf{\scriptsize\textuparrow 1.0}}\\
                                            \textbf{MAD+MADC}  &          &  85.2{\textbf{\scriptsize\textuparrow 9.2}}        & \textbf{74.8} {\textbf{\scriptsize\textuparrow 9.6}}   &90.2 {\textbf{\scriptsize\textuparrow 3.0}}&& 94.4 {\textbf{\scriptsize\textuparrow 0.4}}    & 66.0 {\textbf{\scriptsize\textuparrow 0.8}}   &\textbf{80.8} {\textbf{\scriptsize\textuparrow 1.0}}\\ 
       \midrule
       Single CoT        &\multirow{5}{*}{\parbox{1.7cm}{DeepSeek-R1-Distill-Qwen-14B}}    & 92.4        &    75.2      &87.2\\
                                            SC  &                &99.2 &79.2&  90.8\\
                                           MAD    &              &99.6 &79.6& 90.2\\
                                            MAD+Truth  &  &99.6 \textbf{- -} &\textbf{81.6} {\textbf{\scriptsize\textuparrow 2.0}} &92.2 {\textbf{\scriptsize\textuparrow 2.0}}\\
                                             \textbf{MAD+MADC}   &        &99.6 \textbf{- -} &\textbf{80.4} {\textbf{\scriptsize\textuparrow 0.8}} &90.8 {\textbf{\scriptsize\textuparrow 0.6}}\\  
                              
        \bottomrule
    \end{tabular}
    }
\end{table*}

As shown in Table~\ref{tab:madc_2}, our MADC strategy achieves consistent improvements across major model families. For \textbf{Qwen} series, MADC enhances accuracy by +8.8\% (3B) and +3.6\% (7B) on Logical Deduction, with \textbf{DeepSeek-Qwen-7B} showing +9.6\% improvement on Geometric Shapes. The \textbf{GLM} family attains +3.6\% (Flash) and +1.2\% (FlashX) gains on logic tasks, while \textbf{GPT-4o-mini} maintains +2.0\% improvement.

To explore the potential for improvement in MADC, we implement the Truth Last strategy as an idealized reference. Experimental results show significant room for improvement: Truth Last achieves remarkable gains of 22.8\% on both Logical Deduction and Geometric Shapes for Qwen-2.5-3B-Instruct, and 24.4\% on Geometric Shapes for DeepSeek-R1-Qwen-7B. Even high-performance models like GPT-4o-mini show additional improvements of 3.2\%–4.0\% through this strategy. Notably, the DeepSeek-R1-Qwen-14B model reaches 99.6\% accuracy on Logical Deduction under Correct Last, highlighting the significant room for improvement in black-box implementations.

\subsection{Scaling Experiment (RQ2)}
\begin{figure}[htp]

  \includegraphics[width=\columnwidth]{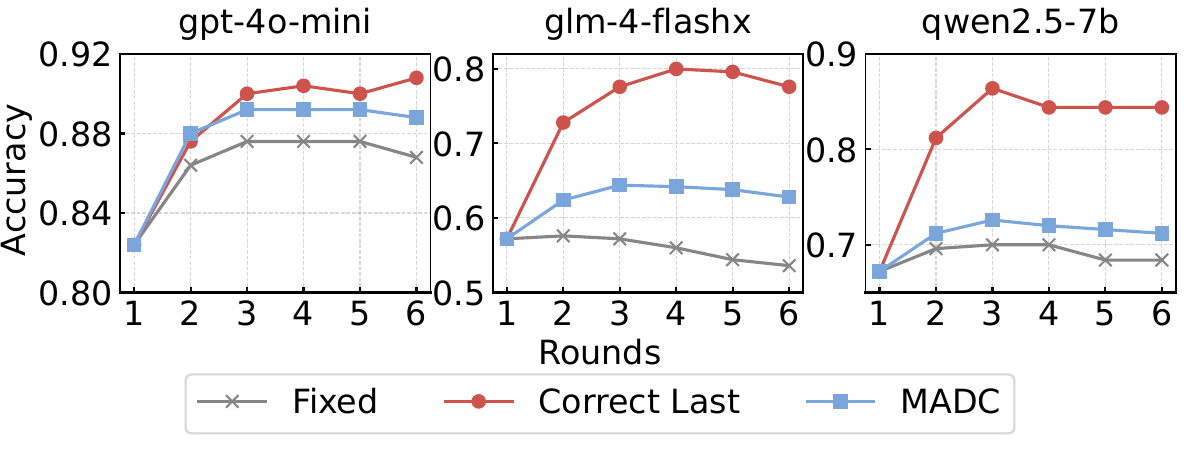} 
  
  \caption {Performance of round scaling.}
  \label{fig:round}
\end{figure}
In the context of MAD frameworks, understanding debate round scaling is crucial to validate our conclusions and methods. We examined whether allocation strategies remain effective and consistent as the number of debate rounds increases. Figure \ref{fig:round} shows that as debate rounds scale, the "Truth Last" strategy significantly outperforms the standard MAD approach. Additionally, the MADC method remains effective as the round scale increases. At the same time, we also validated the performance of MADC as the number of agents scales. The experimental results indicate that MADC remains effective for different numbers of agents, and MADC‘s effectiveness gradually becomes more significant than MAD as the number of agents increases. Detailed experimental results can be found in the Appendix A.3.

\subsection{Experiment on Agent Requirements for Reversing Debate Outcome (RQ3)}
\begin{figure}[hbt]
  \includegraphics[width=1\linewidth]{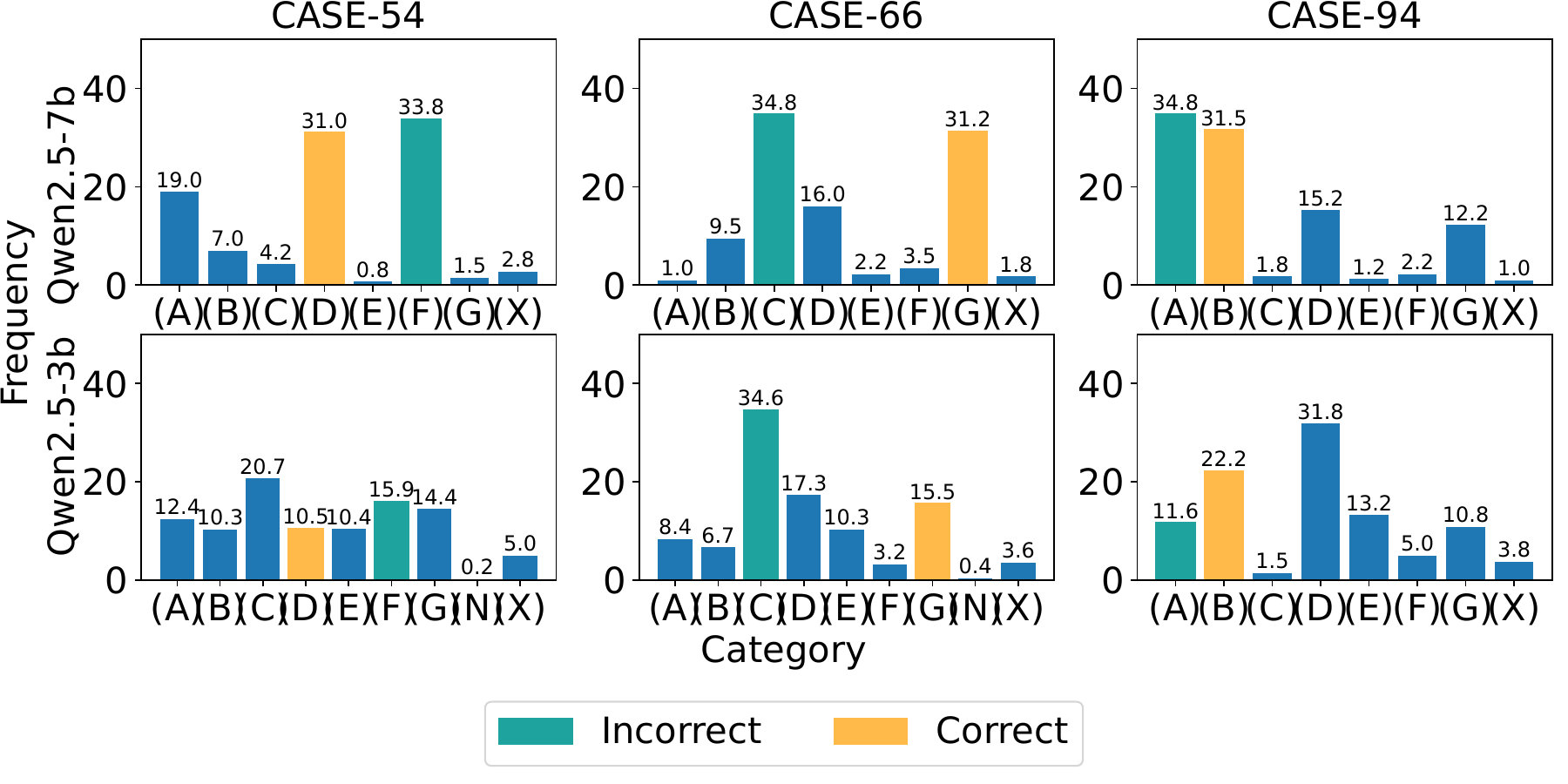} 
  
  \caption {After resampling answers 400 times per question using CoT, we deliberately selected correct and incorrect viewpoints with similar distributions in the 7B model. However, for the same questions in the 3B model, the distributions differ.}
 \label{fig:case_1}
\end{figure}

Based on the experimental setup in Section \ref{sub:exp}, we sampled each question 400 times. For the 7B model, we selected three cases with a similar distribution of correct and incorrect viewpoints, which is shown in Figure \ref{fig:case_1}. 

To explore the relationship between model preference and position preference, we repeated the same experiment using Qwen-2.5-3B-Instruct. It is important to note that, in the 3B model, the distribution of correct and incorrect viewpoints was not similar for these cases.

\begin{figure}[hbt]
  \includegraphics[width=1\columnwidth]{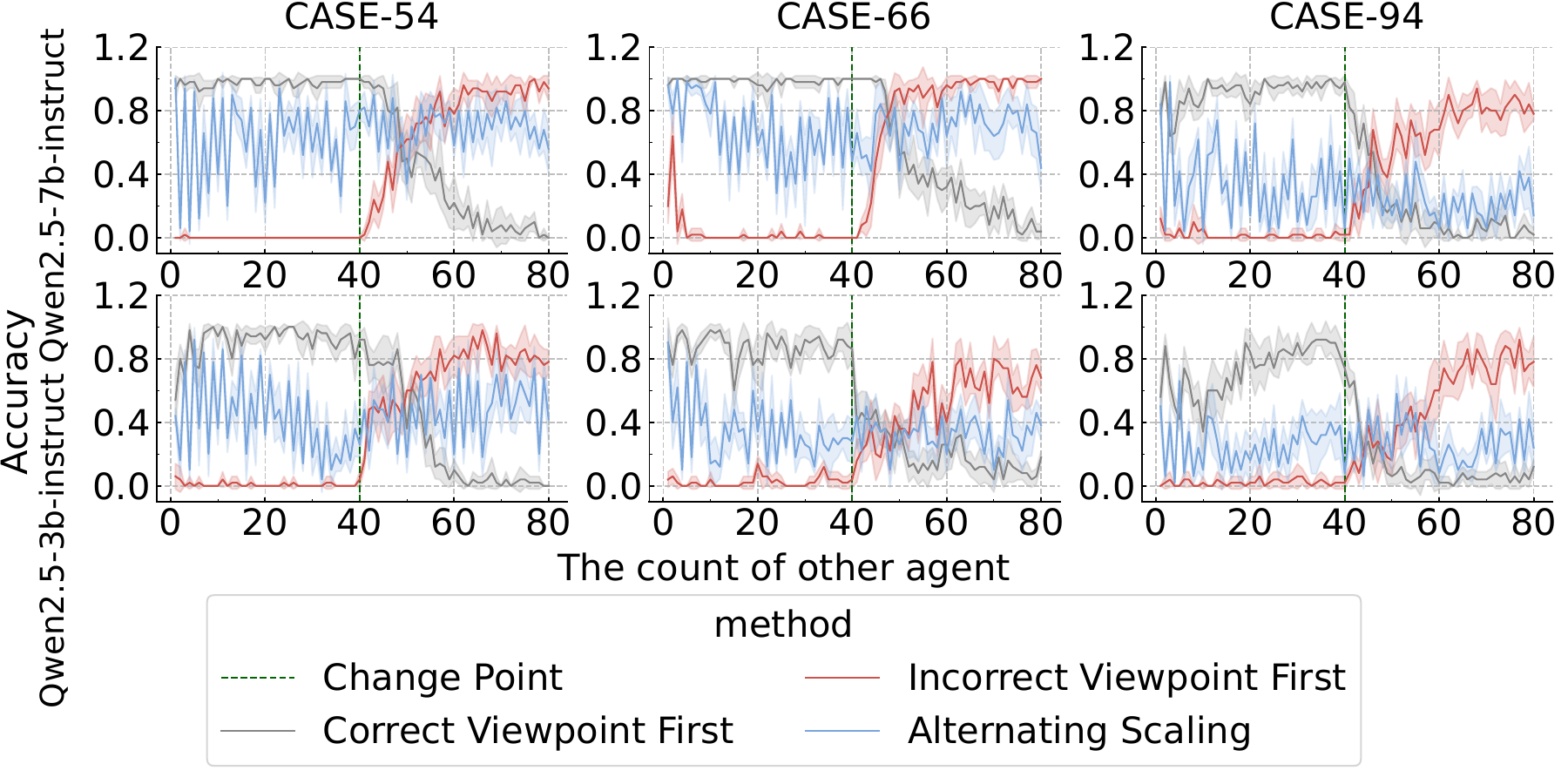} 
  
  \caption {Accuracy variation curves of agent count scaling.}
    \label{fig:case_0}
\end{figure}

Figure \ref{fig:case_0} presents the experimental results for different scaling methods. From this, we can conclude the following:

(1) Allocation strategies significantly impact the MAD scaling law. This is evident as the different strategies do not converge to the same point when scaled to 80 agents.

(2) Agents positioned later have a greater influence on the debate outcome. Given the same context, the results tend to favor the viewpoints of agents speaking later, regardless of whether their viewpoints are correct.

(3) Reverse a conclusion is considerably easier than upholding one. With just 20 agents, we can achieve a 50\% change, even with a base of 40 agents. For detailed data, refer to the Appendix A.6.

When examining the `alternating scale method', we observe that it consistently falls between the other two methods as the number of agents increases to 80, aligning with our intuition. However, the debate outcomes do not stabilize as the number of agents increases. Additionally, there is significant fluctuation when the agent count is low, indicating that `alternating ' is a very bad allocation strategy. Furthermore, even though these viewpoints have different distributions in the Qwen2.5-3B-Instruct, the conclusions we previously reached still hold.

\section{Conclusion}
In our study, we examined the impact of role allocation strategies within multi-agent framework and scaling laws, thoroughly validating our conclusions. Through experiments on nine 9 models, we discovered that role allocation is a critical factor influencing scaling laws in MA. Allocating the role with the correct viewpoint last enhances MAD effectiveness, and an orderly role arrangement facilitates reaching agreement in debates. We propose the MADC method, an orthogonal approach that significantly improves MAD outcomes and is widely applicable within the multi-agent framework. Future research can further explore these strategies, unlocking more potential in MA scaling laws.

\bibliography{main}
\onecolumn

\section{Appendix}
\subsection{Log-likelihood Calculation Example}\label{sec:log-example}
We calculate Log-likelihood as common:

\begin{equation}
    \begin{aligned}
        LL=-\sum_{i=1}^{n}t_ilog_2P(y_{i}) 
    \end{aligned}
\end{equation}
where $t_i=1$ if answer is correct, otherwise 0, We can think of this as a binary classification problem.

A higher log-likelihood represents a higher probability of $P(y_i = \text{Truth})$, which indicates that a greater number of agents are correct.

Let's look a simple described scenario (10 agents, stop after 1 debate round, ground truth is A):

\renewcommand{\arraystretch}{1.5} % 增加行高（1.5 倍）
\begin{center}
\begin{tabular}{lccccc}
\toprule
\textbf{Strategy} & \textbf{Initial Round} & \textbf{Debate Round} & \textbf{Final Answer} & \boldmath$P(y_i=A)$ & \textbf{Log-likelihood} \\ 
\midrule 
Fixed & (A)x5, (B)x5 & (A)x9, (B)x1 & (A) & 0.9 & -0.152 \\ 
Random & (A)x5, (B)x5 & (A)x10 & (A) & 1.0 & 0.0 \\ 
\bottomrule 
\end{tabular}
\end{center}

In this scenario, after a round of debate under the \textbf{Fixed strategy}, not all agents reach  agreement, yet the correct answer can still be derived. However, two strategies have differing log-likelihood metrics. Under resource constraints (e.g., limited debate rounds), when achieving the same number of correct questions, strategies with higher LL demonstrate greater robustness.

\subsection {Entropy Calculation Example}\label{sec:entropy-example}
We calculate Entropy as common:
\begin{equation}
    \begin{aligned}
        H(X)=-\sum_{i=1}^np_ilog_2p_i
    \end{aligned}
\end{equation}
where n represents the total number of viewpoints that have emerged. Lower entropy means the views in the MA system are more concentrated. It's easier to reach an agreement.

Let's look a simple described scenario (10 agents, stop after 1 debate round, ground truth is A):
\begin{center}
\begin{tabular}{lccc}
\toprule 
\textbf{Strategy} & \textbf{Initial Round} & \textbf{Debate Round} & \textbf{Entropy} \\ 
\midrule 
Fixed & (A)x5, (B)x5 & (A)x8, (B)x1, (C)x1 & $-(-0.2575-0.3322-0.3322)=0.9219$ \\ 
Random & (A)x5, (B)x5 & (A)x10 & $0.0$ \\ 
\bottomrule 
\end{tabular} 
\end{center}

A lower entropy indicates a more concentrated viewpoint of debate.
\clearpage

\subsection{Agent Count Scaling Experiment}\label{sec:count_scale}
\begin{table}[h]
\centering 
\caption{Qwen2.5-7B-Instruct / Geometric Shapes}
\begin{tabular}{lcccccccc}
\toprule 
Number of Agents & 3 & 4 & 5 & 6 & 7 & 8 & 9 & 10 \\
\midrule 
MAD & 59.2 & 61.6 & 60.8 & 62.0 & 62.0 & 64.0 & 63.2 & 64.8 \\
MAD+MADC & 58.0 & 64.0 & 62.4 & 64.4 & 62.8 & 64.8 & 67.2 & 68.4 \\
\bottomrule 
\end{tabular}
\end{table}
 
\begin{table}[h]
\centering 
\caption{Qwen2.5-7B-Instruct / Logical Deduction}
\begin{tabular}{lcccccccc}
\toprule 
Number of Agents & 3 & 4 & 5 & 6 & 7 & 8 & 9 & 10 \\
\midrule 
MAD & 61.2 & 62.8 & 65.2 & 66.8 & 67.2 & 67.6 & 67.6 & 70.4 \\
MAD+MADC & 61.6 & 64.8 & 66.0 & 69.2 & 69.6 & 68.0 & 70.0 & 72.0 \\
\bottomrule 
\end{tabular}
\end{table}
 
\begin{table}[h]
\centering 
\caption{DeepSeek-R1-Distill-Qwen-7B / Geometric Shapes}
\begin{tabular}{lcccccccc}
\toprule 
Number of Agents & 3 & 4 & 5 & 6 & 7 & 8 & 9 & 10 \\
\midrule 
MAD & 60 & 61.6 & 62.4 & 64 & 63.2 & 64 & 66 & 65.2 \\
MADC & 61.6 & 63.2 & 66.4 & 71.2 & 70.4 & 74.4 & 74 & 74.8 \\
\bottomrule 
\end{tabular}
\end{table}
 
\begin{table}[!h]
\centering 
\caption{DeepSeek-R1-Distill-Qwen-7B / Logical Deduction}
\begin{tabular}{lcccccccc}
\toprule 
Number of Agents & 3 & 4 & 5 & 6 & 7 & 8 & 9 & 10 \\
\midrule 
MAD & 72 & 72.4 & 73.2 & 74.8 & 73.6 & 76 & 77.2 & 76 \\
MAD+MADC & 74.8 & 75.6 & 78 & 81.2 & 80.8 & 81.2 & 84.4 & 85.2 \\
\bottomrule 
\end{tabular}
\end{table}
\clearpage
\subsection{Example of MADC in MAD}\label{sec:madc_example}
\begin{tcolorbox}[width=1\textwidth,title={\textbf{Fewshot Prompt}}]
		\#\#\# Instruction:
Gracie and Joe are choosing numbers on the complex plane. Joe chooses the point $1+2i$. Gracie chooses $-1+i$. How far apart are Gracie and Joe's points?\\

\#\#\# Response:
Let's think step by step.
Step 1: The distance between two points $(x_1,y_1)$ and $(x_2,y_2)$ in the complex plane is given by the formula $\sqrt{(x_2-x_1)^2+(y_2-y_1)^2}$.
Step 2: In this case, Joe's point is $(1,2)$ and Gracie's point is $(-1,1)$.
Step 3: The distance between their points is $\sqrt{((-1)-(1))^2+((1)-(2))^2}=\sqrt{(-2)^2+(-1)^2}=\sqrt{4+1}=\sqrt{5}$.
Step 4: Therefore, Gracie and Joe's points are $(\sqrt{5})$ units apart.
Step 5: The answer is: $(\sqrt{5})$.\\

\#\#\# Instruction:
What is the sum of all positive integer values of $n$ for which $\frac{n+6}{n}$ is an integer?\\

\#\#\# Response:
Let's think step by step.
Step 1: We want $\frac{n+6}{n}$ to be an integer. This means that $n+6$ must be divisible by $n$.
Step 2: Since $n$ is a positive integer, $n+6$ must be greater than $n$.
Step 3: Therefore, $n$ must be a factor of 6.
Step 4: The positive factors of 6 are 1, 2, 3, and 6. The sum of these factors is $1+2+3+6=(12)$.
Step 5: The answer is: (12).\\

\#\#\# Instruction:
Abigail, Beatrice, and Carson want to sell their eggs at the market. Abigail has 37 eggs, Beatrice has 49 eggs, and Carson has 14 eggs. They can only sell the eggs in cartons of 12. How many eggs will be remaining if they sell all the cartons?\\

\#\#\# Response:
Let's think step by step.
Step 1: First, we find the total number of eggs they have: $37 + 49 + 14 = 100$.
Step 2: Next, we divide the total number of eggs by the number of eggs per carton: $100 \div 12 = 8$ with a remainder of $4$.
Step 3: Therefore, if they sell all the cartons, there will be $(4)$ eggs remaining.
Step 4: The answer is: (4).\\

\#\#\# Instruction:
Circle $T$ has its center at point $T(-2,6)$. Circle $T$ is reflected across the $y$-axis and then translated 8 units down. What are the coordinates of the image of the center of circle $T$?\\

\#\#\# Response:
Let's think step by step.
Step 1: Reflecting a point across the $y$-axis negates its $x$-coordinate.
Step 2: So, the image of point $T(-2,6)$ after reflecting across the $y$-axis is $(-(-2),6)=(2,6)$.
Step 3: Then, translating a point down 8 units subtracts 8 from its $y$-coordinate.
Step 4: So, the image of point $(2,6)$ after translating down 8 units is $(2,6-8)=(2,-2)$. Therefore, the coordinates of the image of the center of circle $T$ are $((2,-2))$.
Step 5: The answer is: ((2,-2)).\\
        
\end{tcolorbox}
~\\\\\\\\\\\\\\\\\\
\begin{center}
	\begin{tcolorbox}[width=1\textwidth,title={\textbf{Round.1 Initial Round}}]
		\#\#\# Agent Input: Can you answer the following question as accurately as possible? At the national curling championships, there are three teams of four players each. After the championships are over, the very courteous participants each shake hands three times with every member of the opposing teams, and once with each member of their own team. How many handshakes are there in total? Explain your answer. \\Make sure putting the answer in the form (X) at the end of your response.\\
        
        \#\#\# Agent Output: Let's think step by step.\\ Step 1: We have three teams, each with four players. The total number of players is \(3 \times 4 = 12\).\\\\Step 2: Each player shakes hands with every member of the opposing teams three times. Since there are two opposing teams, each player will shake hands with \(8\) players (4 players from each of the two opposing teams).\\\\Step 3: For each of the 4 players from one team, the number of handshakes they make with the other teams is: \( 3 \text{ (handshakes)} \times 8 \text{ (opponents)} = 24 \text{ handshakes per player}. \)\\\\Step 4: Since there are 4 players per team, the total handshakes from one team with members of the opposing teams:\( 4 \text{ (players)} \times 24 \text{ (handshakes per player)} = 96 \text{ handshakes per team}.\)\\\\Step 5: Since there are three teams, the total number of handshakes between teams is:\( \\96 \text{ (handshakes per team)} \times 3 \text{ (teams)} = 288 \text{ handshakes between the teams}.\)\\\\Step 6: In addition, each player shakes hands once with each of their own team members. Since there are 4 players on a team and each player shakes hands with the other 3, the total number of handshakes within one team is:\( \text{ handshakes in one team} = \frac{4 \times 3}{2} = 6 \text{ (handshakes)}. \)\\\\Step 7: Since there are 3 teams, the total number of handshakes within all teams is:\( 6 \text{ (handshakes per team)} \times 3 \text{ (teams)} = 18 \text{ handshakes}.\)\\\\Step 8: Finally, we add the handshakes between the teams (288) to the handshakes within the teams (18):\( \text{Total handshakes} = 288 + 18 = 306.\)\\\\Step 9: \textbf{Therefore, the total number of handshakes is 306.The answer is: (306).}\\\\

	\end{tcolorbox}
\end{center}

\begin{center}
	\small
    \scalebox{0.85}{
	\begin{tcolorbox}[width=1\textwidth,title={\textbf{Round.2 Debate Round-1}}]
		\#\#\# Agent Input: These are the solutions to the problem from other agents: \\\\ One agent solution: \\ Step 1: First, we identify the number of teams and players. There are 3 teams with 4 players each, giving us a total of \(3 \times 4 = 12\) players.\\\\ Step 2: Each player shakes hands 3 times with every member of the opposing teams. Since there are 2 opposing teams (each with 4 players), each player will have \(4 \times 2 = 8\) opposing players to shake hands with.\\\\ Step 3: The total number of handshakes between each player and the opposing players is \(3 \times 8 = 24\) handshakes per player.\\\\Step 4: Since there are 12 players in total, the total number of handshakes from the players shaking hands with opposing players is \(12 \times 24 = 288\).\\\\ Step 5: Additionally, each player shakes hands once with every member of their own team. With 4 players per team, each player shakes hands with the other 3 players on their team.\\\\ Step 6: The total number of in-team handshakes for each team is \(3\) handshakes per player for \(4\) players, yielding \(4 \times 3 = 12\) handshakes per team.\\\\ Step 7: Since there are 3 teams, the total number of in-team handshakes is \(3 \times 12 = 36\).\\\\ Step 8: Therefore, the total number of handshakes, combining both in-team and opposing team handshakes, is \(288 + 36 = 324\).\\\\ Step 9: \textbf{The final answer is: (324)}\\\\

        One agent solution: \\ Step 1: We have three teams, each with four players. The total number of players is \(3 \times 4 = 12\).\\\\Step 2: Each player shakes hands with every member of the opposing teams three times. Since there are two opposing teams, each player will shake hands with \(8\) players (4 players from each of the two opposing teams).\\\\Step 3: For each of the 4 players from one team, the number of handshakes they make with the other teams is: \(3 \text{ (handshakes)} \times 8 \text{ (opponents)} = 24 \text{ handshakes per player}. \)\\\\Step 4: Since there are 4 players per team, the total handshakes from one team with members of the opposing teams:\(4 \text{ (players)} \times 24 \text{ (handshakes per player)} = 96 \text{ handshakes per team}.\)\\\\Step 5: Since there are three teams, the total number of handshakes between teams is:\(96 \text{ (handshakes per team)} \times 3 \text{ (teams)} = 288 \text{ handshakes between the teams}.\)\\\\Step 6: In addition, each player shakes hands once with each of their own team members. Since there are 4 players on a team and each player shakes hands with the other 3, the total number of handshakes within one team is:\(\text{ handshakes in one team} = \frac{4 \times 3}{2} = 6 \text{ (handshakes)}. \)\\\\Step 7: Since there are 3 teams, the total number of handshakes within all teams is:\(6 \text{ (handshakes per team)} \times 3 \text{ (teams)} = 18 \text{ handshakes}.\)\\\\Step 8: Finally, we add the handshakes between the teams (288) to the handshakes within the teams (18):\(\text{Total handshakes} = 288 + 18 = 306.\)\\\\Step 9: Therefore, the total number of handshakes is 306. \textbf{The answer is: (306).}\\\\ 
        
        One agent solution: \\Step 1: First, we determine the number of teams and players. There are 3 teams, each with 4 players. Therefore, the total number of players is \(3 \times 4 = 12\).\\\\Step 2: Each player shakes hands three times with every member of the opposing teams. Since each team has 4 players, each player's opposing players from the other two teams are \(4 + 4 = 8\) players.\\\\Step 3: Each player shakes hands 3 times with these 8 opposing players, resulting in a total of handshakes from one player to the opposing players: \(8 \times 3 = 24 \text{ handshakes per player}.\)
	
	\end{tcolorbox}
    }
\end{center}

\begin{center}
	\small
    \scalebox{0.95}{
	\begin{tcolorbox}[width=1\textwidth,title={\textbf{Round.2 Debate Round-2}}]
	Step 4: With 12 players in total, the total handshakes among the opposing players is:\(12 \times 24 = 288 \text{ handshakes}. \)\\\\Step 5: Next, each player shakes hands once with each member of their own team (3 other members). Therefore, each player contributes:\(3 \text{ handshakes with their own team}.\)\\\\Step 6: For 12 players, the total handshakes within their own teams are:\(12 \times 3 = 36 \text{ handshakes}. \)\\\\Step 7: However, because the handshakes within a team are counted for each player, we have counted them too many times. Since each handshake among team members is counted twice (once for each participant), we need to divide this by 2:\(\frac{36}{2} = 18 \text{ distinct handshakes within teams}.\)\\\\Step 8: Finally, we sum the total handshakes:\(288 + 18 = 306.\)\\\\Step 9: Thus, the total number of handshakes that occur is:\(\text{The answer is: } 306.\) \textbf{In conclusion, the total number of handshakes is (306).}\\\\ 
        
        One agent solution: \\Step 1: There are three teams, each with 4 players. Therefore, the total number of players is \(3 \times 4 = 12\).\\\\Step 2: Each player shakes hands three times with every player on the opposing teams. Since there are two opposing teams, each player has \(4 + 4 = 8\) players to shake hands with on the opposing teams. Thus, each player shakes hands \(3 \times 8 = 24\) times with players from the opposing teams.\\\\ Step 3: Now, each player shakes hands once with each member of their own team, which consists of 3 other players. Thus, each player shakes hands \(3 \times 1 = 3\) times with their own team.\\\\ Step 4: Therefore, the total handshakes for each player is \(24 + 3 = 27\) handshakes.\\\\ Step 5: Since there are 12 players, the total handshakes can be calculated as \(27 \times 12 = 324\).\\\\ Step 6: However, since every handshake is counted twice in our previous calculations (once for each participant in the handshake), we need to divide the total by 2. Thus, the total number of unique handshakes is \(\frac{324}{2} = 162\).  \textbf{Final Answer: The total number of handshakes is (162).}\\\\

        One agent solutions: \\Step 1: First, let's determine the number of teams and players. There are 3 teams, each consisting of 4 players. Thus, the total number of players is:\(3 \text{ teams} \times 4 \text{ players/team} = 12 \text{ players.}\)\\\\Step 2: Now, we calculate the number of handshakes each player will make. Each player shakes hands 3 times with every member of the opposing teams and once with each member of their own team.\\\\Step 3: A player has 8 opponents (4 players from each of the 2 opposing teams) and will shake hands 3 times with each of them. Hence, the number of handshakes with opponents for one player is:\(8 \text{ opponents} \times 3 \text{ handshakes/opponent} = 24 \text{ handshakes.}\)\\\\Step 4: Each player also shakes hands once with each member of their own team. Since there are 3 teammates (4 total players minus themselves), the number of handshakes with teammates is:\(3 \text{ teammates} \times 1 \text{ handshake/team member} = 3 \text{ handshakes.}\)\\\\Step 5: Therefore, the total number of handshakes for one player is:\(24 \text{ (with opponents)} + 3 \text{ (with teammates)} = 27 \text{ handshakes.}\)\\\\Step 6: Since there are 12 players in total, the total number of handshakes (without considering that each handshake is counted twice) is given by:\(12 \text{ players} \times 27 \text{ handshakes/player} = 324 \text{ handshakes.}\)\\\\Step 7: Since each handshake is counted twice (once for each participant), we need to divide this total by 2:\(\frac{324}{2} = 162 \text{ unique handshakes.}\)\\\\Step 8: Therefore, the total number of unique handshakes at the curling championships is \( 162 \). \textbf{The answer is: (162)}\\\\

        One agent solutions:\\ Step 1: There are 3 teams, and each team consists of 4 players. Therefore, the total number of players is \(3 \times 4 = 12\).\\\\Step 2: Each player shakes hands with every member of the opposing teams three times and shakes hands once with each member of their own team.

	\end{tcolorbox}
    }
\end{center}

\begin{center}
	\small
	\begin{tcolorbox}[width=1\textwidth,title={\textbf{Round.2 Debate Round-3}}]
	Step 3: Each team has 4 players, so for a given player, they will shake hands once with each of the 3 other members of their own team. Thus, a player shakes hands \(3\) times with their own team.\\\\Step 4: The number of opposing players for a given player, since there are \(2\) opposing teams each with \(4\) players, is \(8\).\\\\Step 5: Each player shakes hands \(3\) times with each member of the opposing teams. Therefore, the number of handshakes with opposing players per player is \(8 \times 3 = 24\).\\\\Step 6: The total number of handshakes for one player is the sum of handshakes with their own team and the opposing teams: \( 3 \text{ (with own team)} + 24 \text{ (with opposing teams)} = 27. \)\\\\Step 7: There are \(12\) players in total, and since the handshakes are counted for each player, we calculate the total number of handshakes as follows:\(\text{Total handshakes} = 12 \times 27 = 324\).\\\\Step 8: However, this counts each handshake twice (once for each participant). Therefore, we must divide by \(2\):\(\text{Unique handshakes} = \frac{324}{2} = 162\).\\\\Step 9: The total number of unique handshakes is \(162\).\\\\Step 10: \textbf{The answer is: (162).}\\\\

    \textbf{Using the reasoning from other agents as additional advice, can you give an updated answer? Examine your solution and that other agents step by step. Put your answer in the form (X) at the end of your response.}\\\

    \#\#\# Agent Output: Let's analyze and revise my previous solution step by step, incorporating insights from the other agents.\\\\Step 1: There are three teams, each with four players. Therefore, the total number of players is:\(3 \text{ teams} \times 4 \text{ players/team} = 12 \text{ players}\)\\\\Step 2: Each player shakes hands three times with each player from the other teams. Since there are two other teams, each with four players, the total number of players in the other teams is:\(4 \text{ players/team} \times 2 \text{ teams} = 8 \text{ players}\)\\\\Step 3: The number of handshakes each player has with players from the other teams is:\(3 \text{ handshakes/player} \times 8 \text{ players} = 24 \text{ handshakes}\)\\\\Step 4: Since there are 12 players total, the total number of handshakes from this interaction (with the other teams) is:\(12 \text{ players} \times 24 \text{ handshakes} = 288\)\\\\Step 5: Next, each player shakes hands once with each member of their own team. Since there are 3 players on their own team to shake hands with (4 total minus themselves), each player shakes hands:\(3 \text{ handshakes/player} \times 3 \text{ teams} \times 4 \text{ players/team} = 36 \text{ handshakes}\)\\\\Step 6: Now, add the handshakes from both interactions:\(288 \text{ (other teams handshakes)} + 36 \text{ (own team handshakes)} = 324 \text{ total handshakes}\)\\\\Step 7: However, we need to divide the handshakes by 2 because every handshake is counted twice (once for each participant):\(\\frac{324}{2} = 162\)\\\\Step 8: Therefore, the total number of handshakes that occurred at the championship is:\(162\)\\\\Step 9: \textbf{The answer is: (162).}
    
	\end{tcolorbox}
\end{center}

\subsection{Case Detail for RQ3}\label{sec:threshold_case}

\begin{center}
	\small
	\begin{tcolorbox}[width=1\textwidth,title={\textbf{CASE-54}}]
	The following paragraphs each describe a set of seven objects arranged in a fixed order. The statements are logically consistent within each paragraph. In an antique car show, there are seven vehicles: a truck, a motorcyle, a sedan, a minivan, a station wagon, a hatchback, and a tractor. The hatchback is newer than the truck. The sedan is the third-newest. The station wagon is the newest. The motorcyle is older than the truck. The minivan is newer than the sedan. The tractor is the third-oldest.\\Options:\\(A) The truck is the second-newest\\(B) The motorcyle is the second-newest\\(C) The sedan is the second-newest\\(D) The minivan is the second-newest\\(E) The station wagon is the second-newest\\(F) The hatchback is the second-newest\\(G) The tractor is the second-newest

	\end{tcolorbox}
\end{center}
\begin{center}
	\small
	\begin{tcolorbox}[width=1\textwidth,title={\textbf{CASE-66}}]
	The following paragraphs each describe a set of seven objects arranged in a fixed order. The statements are logically consistent within each paragraph. A fruit stand sells seven fruits: watermelons, oranges, mangoes, cantaloupes, kiwis, pears, and peaches. The pears are the second-cheapest. The peaches are more expensive than the cantaloupes. The peaches are less expensive than the mangoes. The cantaloupes are more expensive than the kiwis. The oranges are the fourth-most expensive. The watermelons are the second-most expensive.\\Options:\\(A) The watermelons are the third-most expensive\\(B) The oranges are the third-most expensive\\(C) The mangoes are the third-most expensive\\(D) The cantaloupes are the third-most expensive\\(E) The kiwis are the third-most expensive\\(F) The pears are the third-most expensive\\(G) The peaches are the third-most expensive

	\end{tcolorbox}
\end{center}
\begin{center}
	\small
	\begin{tcolorbox}[width=1\textwidth,title={\textbf{CASE-94}}]
The following paragraphs each describe a set of seven objects arranged in a fixed order. The statements are logically consistent within each paragraph. On a shelf, there are seven books: a green book, a brown book, a white book, a black book, an orange book, a purple book, and a yellow book. The purple book is to the left of the yellow book. The green book is to the left of the white book. The brown book is to the right of the yellow book. The white book is the fourth from the left. The green book is to the right of the orange book. The black book is the second from the left.\\Options:\\(A) The green book is the second from the right\\(B) The brown book is the second from the right\\(C) The white book is the second from the right\\(D) The black book is the second from the right\\(E) The orange book is the second from the right\\(F) The purple book is the second from the right\\(G) The yellow book is the second from the right
	\end{tcolorbox}
\end{center}
\newpage
\subsection{Detailed Experimental Data on Agent Threshold for Reversing Debate Outcomes(RQ3)} 
\label{sec:threshold}
\begin{table*}[!htbp]
    \centering
    \caption{CASE-54 Qwen2.5-7B-Instruct}

    \small
     \scalebox{0.85}{
    \begin{tabular}{cccccccc}
        \toprule
        \textbf{Count} & \textbf{Correct First} & \textbf{Incorrect First}     &\textbf{Alternating Scale} &\textbf{Count} & \textbf{Correct First} & \textbf{Incorrect First}     &\textbf{Alternating Scale}\\ 
        \midrule 
        
        1&0.94&0.00&1.00&41&0.98&0.02&0.82\\2&1.00&0.00&0.06&42&0.98&0.10&0.70\\3&0.96&0.02&0.94&43&0.94&0.24&0.90\\4&0.98&0.00&0.04&44&0.96&0.16&0.54\\5&0.98&0.00&0.92&45&0.94&0.26&0.80\\6&0.92&0.00&0.12&46&0.80&0.48&0.56\\7&0.94&0.00&0.66&47&0.78&0.30&0.88\\8&0.94&0.00&0.28&48&0.82&0.56&0.54\\9&0.94&0.00&0.88&49&0.48&0.58&0.66\\10&1.00&0.00&0.40&50&0.50&0.62&0.34\\11&0.98&0.00&0.88&51&0.38&0.62&0.70\\12&1.00&0.00&0.20&52&0.54&0.72&0.58\\13&0.98&0.00&0.90&53&0.52&0.72&0.84\\14&0.96&0.00&0.78&54&0.50&0.76&0.64\\15&1.00&0.00&0.74&55&0.46&0.72&0.84\\16&1.00&0.00&0.54&56&0.36&0.84&0.76\\17&1.00&0.00&0.82&57&0.44&0.92&0.78\\18&0.96&0.00&0.32&58&0.24&0.70&0.72\\19&0.96&0.00&0.68&59&0.18&0.82&0.80\\20&1.00&0.00&0.22&\textbf{60}&0.22&0.78&0.58\\
21&1.00&0.00&0.78&61&0.16&0.84&0.80\\22&0.96&0.00&0.32&62&0.12&0.96&0.54\\23&0.96&0.00&0.94&63&0.20&0.90&0.80\\24&0.98&0.00&0.64&64&0.06&0.94&0.60\\25&1.00&0.00&0.76&65&0.16&0.94&0.80\\26&0.94&0.00&0.66&66&0.02&0.92&0.62\\27&0.96&0.00&0.82&67&0.12&0.92&0.78\\28&0.98&0.00&0.52&68&0.04&0.86&0.68\\29&1.00&0.00&0.82&69&0.06&0.92&0.82\\30&0.96&0.00&0.54&70&0.08&0.92&0.72\\31&0.94&0.00&0.72&71&0.04&0.92&0.86\\32&0.98&0.00&0.50&72&0.08&0.90&0.68\\33&0.98&0.00&0.86&73&0.04&0.96&0.76\\34&0.98&0.00&0.42&74&0.02&0.96&0.70\\35&0.98&0.00&0.62&75&0.06&0.86&0.80\\36&0.98&0.00&0.26&76&0.08&0.98&0.62\\37&1.00&0.00&0.86&77&0.10&1.00&0.70\\38&1.00&0.00&0.68&78&0.00&0.92&0.56\\39&1.00&0.00&0.70&79&0.02&0.98&0.68\\40&1.00&0.00&0.80&80&0.00&0.94&0.56\\
    \bottomrule
    
    \end{tabular}}
\end{table*}

\begin{table*}[ht b p]
    \centering
    \caption{CASE-66 Qwen2.5-7B-Instruct}

    \small
     \scalebox{0.85}{
    \begin{tabular}{cccccccc}
        \toprule
        \textbf{Count} & \textbf{Correct First} & \textbf{Incorrect First}     &\textbf{Alternating Scale} &\textbf{Count} & \textbf{Correct First} & \textbf{Incorrect First}     &\textbf{Alternating Scale}\\ 
        \midrule 
        
1&0.96&0.20&0.96&41&1.00&0.00&0.66\\2&1.00&0.64&0.78&42&1.00&0.08&0.50\\3&1.00&0.04&1.00&43&1.00&0.10&0.52\\4&1.00&0.18&0.60&44&1.00&0.22&0.42\\5&0.98&0.00&0.98&45&0.98&0.48&0.78\\6&1.00&0.02&0.94&46&1.00&0.64&0.84\\7&1.00&0.02&0.96&47&0.92&0.74&0.76\\8&0.98&0.02&0.94&48&0.68&0.80&0.64\\9&0.98&0.00&0.84&49&0.80&0.92&0.82\\10&1.00&0.00&0.78&50&0.52&0.94&0.42\\11&1.00&0.00&0.98&51&0.54&0.84&0.72\\12&1.00&0.00&0.52&52&0.40&0.94&0.64\\13&1.00&0.00&0.80&53&0.46&0.92&0.80\\14&1.00&0.00&0.40&54&0.34&0.96&0.68\\15&0.98&0.00&0.76&55&0.40&0.88&0.66\\16&1.00&0.00&0.40&56&0.32&0.92&0.50\\17&1.00&0.02&0.82&57&0.44&0.96&0.72\\18&1.00&0.00&0.54&58&0.32&0.82&0.60\\19&0.98&0.00&0.86&59&0.30&0.92&0.88\\20&0.96&0.00&0.58&\textbf{60}&0.32&0.96&0.64\\
21&0.96&0.00&0.68&61&0.38&0.94&0.76\\22&0.92&0.00&0.44&62&0.22&0.98&0.70\\23&0.96&0.02&0.90&63&0.36&0.98&0.92\\24&1.00&0.00&0.28&64&0.20&1.00&0.76\\25&0.96&0.04&0.70&65&0.18&0.96&0.90\\26&1.00&0.00&0.32&66&0.20&0.98&0.72\\27&1.00&0.00&0.54&67&0.20&1.00&0.84\\28&0.98&0.00&0.34&68&0.22&1.00&0.72\\29&0.98&0.04&0.76&69&0.08&0.98&0.68\\30&0.98&0.00&0.36&70&0.20&0.98&0.64\\31&0.98&0.00&0.76&71&0.12&0.92&0.66\\32&0.98&0.00&0.52&72&0.26&1.00&0.72\\33&0.96&0.02&0.68&73&0.12&1.00&0.84\\34&1.00&0.00&0.46&74&0.18&0.94&0.78\\35&1.00&0.00&0.88&75&0.04&1.00&0.80\\36&1.00&0.00&0.52&76&0.02&1.00&0.66\\37&0.96&0.00&0.88&77&0.06&0.98&0.84\\38&1.00&0.00&0.44&78&0.12&0.98&0.68\\39&1.00&0.00&0.82&79&0.04&0.98&0.66\\40&1.00&0.00&0.48&80&0.04&1.00&0.44\\
    \bottomrule
    
    \end{tabular}
    }
\end{table*}

\begin{table*}[htbp]
    \centering
    \caption{CASE-94 Qwen2.5-7B-Instruct}

    \small
     \scalebox{0.85}{
    \begin{tabular}{cccccccc}
        \toprule
        \textbf{Count} & \textbf{Correct First} & \textbf{Incorrect First}     &\textbf{Alternating Scale} &\textbf{Count} & \textbf{Correct First} & \textbf{Incorrect First}     &\textbf{Alternating Scale}\\ 
        \midrule 
        
1&0.78&0.12&0.84&41&0.82&0.02&0.50\\2&0.98&0.02&0.04&42&0.78&0.24&0.22\\3&0.64&0.02&0.98&43&0.56&0.12&0.40\\4&0.66&0.00&0.20&44&0.72&0.30&0.16\\5&0.86&0.06&0.42&45&0.52&0.32&0.44\\6&0.84&0.00&0.04&46&0.54&0.68&0.20\\7&0.94&0.00&0.32&47&0.28&0.52&0.40\\8&0.84&0.10&0.40&48&0.20&0.42&0.14\\9&0.82&0.04&0.62&49&0.16&0.38&0.26\\10&0.88&0.06&0.04&50&0.20&0.52&0.14\\11&1.00&0.00&0.52&51&0.22&0.72&0.24\\12&0.94&0.00&0.54&52&0.10&0.64&0.12\\13&0.94&0.00&0.74&53&0.18&0.50&0.40\\14&0.96&0.00&0.28&54&0.10&0.74&0.14\\15&0.92&0.00&0.36&55&0.22&0.64&0.48\\16&0.94&0.00&0.06&56&0.06&0.66&0.34\\17&0.96&0.00&0.62&57&0.06&0.52&0.22\\18&0.96&0.02&0.20&58&0.06&0.66&0.14\\19&0.88&0.02&0.54&59&0.10&0.68&0.16\\20&0.84&0.00&0.14&\textbf{60}&0.08&0.68&0.06\\
21&0.92&0.00&0.72&61&0.02&0.78&0.28\\22&0.90&0.00&0.10&62&0.06&0.90&0.06\\23&1.00&0.00&0.34&63&0.00&0.80&0.26\\24&0.96&0.02&0.04&64&0.02&0.72&0.16\\25&0.94&0.02&0.32&65&0.00&0.80&0.28\\26&0.96&0.00&0.06&66&0.00&0.82&0.06\\27&0.96&0.00&0.40&67&0.02&0.84&0.28\\28&0.98&0.00&0.24&68&0.12&0.94&0.16\\29&0.94&0.02&0.62&69&0.06&0.78&0.14\\30&0.96&0.02&0.10&70&0.04&0.82&0.06\\31&0.92&0.02&0.28&71&0.12&0.80&0.26\\32&0.98&0.00&0.08&72&0.10&0.80&0.10\\33&0.94&0.00&0.26&73&0.08&0.74&0.28\\34&0.98&0.02&0.26&74&0.14&0.82&0.18\\35&0.94&0.02&0.50&75&0.00&0.86&0.30\\36&0.98&0.00&0.18&76&0.00&0.90&0.14\\37&1.00&0.00&0.58&77&0.04&0.86&0.42\\38&0.96&0.04&0.20&78&0.08&0.78&0.30\\39&0.94&0.02&0.42&79&0.04&0.84&0.38\\40&0.98&0.02&0.06&80&0.02&0.78&0.14\\
    \bottomrule
    
    \end{tabular}
    }
\end{table*}

\begin{table*}[htbp]
    \centering
    \caption{CASE-54 Qwen2.5-3B-Instruct}

    \small
         \scalebox{0.85}{
    \begin{tabular}{cccccccc}
        \toprule
        \textbf{Count} & \textbf{Correct First} & \textbf{Incorrect First}     &\textbf{Alternating Scale} &\textbf{Count} & \textbf{Correct First} & \textbf{Incorrect First}     &\textbf{Alternating Scale}\\ 
        \midrule 
        
1&0.54&0.06&0.44&41&0.92&0.16&0.48\\2&0.80&0.04&0.16&42&0.76&0.48&0.22\\3&0.68&0.00&0.78&43&0.76&0.50&0.54\\4&0.98&0.02&0.10&44&0.78&0.46&0.48\\5&0.76&0.00&0.92&45&0.76&0.56&0.48\\6&0.74&0.02&0.36&46&0.78&0.40&0.40\\7&0.96&0.00&0.84&47&0.76&0.54&0.70\\8&0.98&0.00&0.20&48&0.86&0.50&0.20\\9&1.00&0.00&0.78&49&0.64&0.44&0.52\\10&0.94&0.00&0.36&50&0.36&0.60&0.18\\11&1.00&0.04&0.86&51&0.62&0.60&0.52\\12&0.92&0.00&0.30&52&0.56&0.72&0.36\\13&0.80&0.02&0.56&53&0.46&0.66&0.70\\14&0.98&0.02&0.44&54&0.24&0.68&0.32\\15&0.88&0.00&0.82&55&0.32&0.72&0.68\\16&0.96&0.00&0.20&56&0.12&0.66&0.38\\17&0.94&0.02&0.58&57&0.14&0.86&0.48\\18&0.94&0.00&0.36&58&0.12&0.72&0.30\\19&0.96&0.00&0.82&59&0.04&0.78&0.60\\20&0.92&0.00&0.32&\textbf{60}&0.10&0.82&0.20\\
21&0.96&0.00&0.70&61&0.06&0.80&0.74\\22&1.00&0.00&0.42&62&0.02&0.86&0.24\\23&0.94&0.00&0.48&63&0.00&0.80&0.70\\24&0.98&0.02&0.28&64&0.04&0.94&0.34\\25&1.00&0.00&0.66&65&0.02&0.86&0.66\\26&1.00&0.02&0.22&66&0.02&0.98&0.18\\27&0.94&0.00&0.56&67&0.04&0.88&0.46\\28&0.92&0.00&0.14&68&0.08&0.76&0.52\\29&0.96&0.00&0.60&69&0.02&0.96&0.70\\30&0.86&0.00&0.18&70&0.00&0.72&0.50\\31&0.98&0.00&0.36&71&0.04&0.82&0.70\\32&0.96&0.00&0.04&72&0.00&0.80&0.42\\33&0.96&0.02&0.38&73&0.02&0.76&0.52\\34&0.90&0.02&0.10&74&0.04&0.84&0.66\\35&0.86&0.00&0.20&75&0.04&0.86&0.58\\36&0.92&0.00&0.14&76&0.02&0.78&0.50\\37&0.94&0.00&0.32&77&0.02&0.82&0.74\\38&0.96&0.00&0.16&78&0.02&0.80&0.20\\39&0.80&0.00&0.36&79&0.00&0.76&0.68\\40&0.92&0.04&0.26&80&0.00&0.78&0.42\\
    \bottomrule
    
    \end{tabular}
    }
\end{table*}

\begin{table*}[htbp]
    \centering
    \caption{CASE-66 Qwen2.5-3B-Instruct}

    \small
         \scalebox{0.85}{
    \begin{tabular}{cccccccc}
        \toprule
        \textbf{Count} & \textbf{Correct First} & \textbf{Incorrect First}     &\textbf{Alternating Scale} &\textbf{Count} & \textbf{Correct First} & \textbf{Incorrect First}     &\textbf{Alternating Scale}\\ 
        \midrule 
        
1&0.90&0.04&0.90&41&0.40&0.16&0.44\\2&0.76&0.06&0.40&42&0.46&0.22&0.22\\3&0.94&0.02&0.62&43&0.42&0.26&0.50\\4&0.98&0.00&0.10&44&0.42&0.38&0.34\\5&0.92&0.00&0.78&45&0.48&0.20&0.40\\6&0.76&0.02&0.18&46&0.34&0.18&0.34\\7&0.86&0.02&0.76&47&0.36&0.34&0.36\\8&0.90&0.00&0.38&48&0.32&0.22&0.10\\9&0.94&0.00&0.52&49&0.24&0.44&0.34\\10&0.98&0.04&0.14&50&0.38&0.36&0.26\\11&0.96&0.00&0.16&51&0.24&0.36&0.36\\12&0.98&0.00&0.12&52&0.12&0.32&0.30\\13&0.90&0.00&0.40&53&0.14&0.60&0.36\\14&0.90&0.00&0.28&54&0.08&0.48&0.34\\15&0.60&0.00&0.48&55&0.14&0.62&0.50\\16&0.80&0.00&0.34&56&0.10&0.58&0.26\\17&0.90&0.02&0.36&57&0.16&0.78&0.28\\18&0.92&0.02&0.14&58&0.10&0.32&0.24\\19&0.76&0.02&0.60&59&0.26&0.52&0.44\\20&0.80&0.14&0.34&\textbf{60}&0.26&0.40&0.20\\
21&0.76&0.06&0.54&61&0.18&0.54&0.36\\22&0.96&0.06&0.12&62&0.30&0.76&0.34\\23&0.92&0.02&0.48&63&0.32&0.80&0.48\\24&0.86&0.04&0.14&64&0.16&0.62&0.30\\25&0.94&0.02&0.46&65&0.14&0.74&0.40\\26&0.88&0.00&0.24&66&0.10&0.62&0.44\\27&0.74&0.00&0.54&67&0.12&0.60&0.34\\28&0.88&0.00&0.18&68&0.12&0.72&0.20\\29&0.82&0.00&0.32&69&0.08&0.52&0.40\\30&0.92&0.00&0.26&70&0.04&0.80&0.10\\31&0.90&0.02&0.24&71&0.18&0.78&0.24\\32&0.92&0.02&0.12&72&0.04&0.74&0.18\\33&0.94&0.06&0.38&73&0.16&0.74&0.54\\34&0.86&0.12&0.26&74&0.12&0.58&0.32\\35&0.80&0.04&0.22&75&0.10&0.62&0.30\\36&0.82&0.04&0.30&76&0.08&0.56&0.22\\37&0.92&0.04&0.26&77&0.08&0.56&0.38\\38&0.92&0.02&0.30&78&0.04&0.68&0.32\\39&0.88&0.02&0.30&79&0.06&0.76&0.46\\40&0.86&0.04&0.28&80&0.18&0.68&0.38\\
    \bottomrule
    
    \end{tabular}
    }
\end{table*}

\begin{table*}[h b p tt]
    \centering
    \caption{CASE-94 Qwen2.5-3B-Instruct}

    \small
         \scalebox{0.85}{
    \begin{tabular}{cccccccc}
        \toprule
        \textbf{Count} & \textbf{Correct First} & \textbf{Incorrect First}     &\textbf{Alternating Scale} &\textbf{Count} & \textbf{Correct First} & \textbf{Incorrect First}     &\textbf{Alternating Scale}\\ 
        \midrule 
        
1&0.56&0.00&0.50&41&0.62&0.08&0.34\\2&0.88&0.02&0.06&42&0.66&0.16&0.16\\3&0.64&0.04&0.40&43&0.40&0.08&0.40\\4&0.50&0.00&0.02&44&0.22&0.20&0.06\\5&0.42&0.00&0.66&45&0.28&0.38&0.48\\6&0.74&0.04&0.08&46&0.14&0.24&0.28\\7&0.66&0.04&0.34&47&0.26&0.28&0.44\\8&0.50&0.00&0.06&48&0.18&0.24&0.16\\9&0.54&0.00&0.24&49&0.08&0.18&0.34\\10&0.34&0.02&0.10&50&0.06&0.16&0.16\\11&0.60&0.00&0.44&51&0.08&0.38&0.58\\12&0.60&0.00&0.42&52&0.12&0.38&0.32\\13&0.60&0.04&0.28&53&0.04&0.42&0.46\\14&0.54&0.02&0.10&54&0.02&0.50&0.34\\15&0.68&0.00&0.28&55&0.04&0.34&0.46\\16&0.60&0.02&0.08&56&0.06&0.46&0.12\\17&0.78&0.02&0.22&57&0.10&0.42&0.40\\18&0.64&0.02&0.10&58&0.10&0.50&0.24\\19&0.90&0.00&0.22&59&0.08&0.60&0.22\\20&0.84&0.02&0.16&\textbf{60}&0.02&0.60&0.08\\
21&0.74&0.02&0.26&61&0.02&0.74&0.18\\22&0.76&0.06&0.14&62&0.02&0.68&0.26\\23&0.76&0.00&0.30&63&0.00&0.64&0.20\\24&0.84&0.02&0.16&64&0.04&0.72&0.10\\25&0.74&0.04&0.36&65&0.08&0.76&0.16\\26&0.84&0.04&0.12&66&0.04&0.86&0.12\\27&0.86&0.02&0.26&67&0.06&0.70&0.12\\28&0.84&0.02&0.14&68&0.04&0.86&0.20\\29&0.82&0.04&0.32&69&0.04&0.80&0.36\\30&0.90&0.06&0.26&70&0.04&0.74&0.20\\31&0.82&0.06&0.44&71&0.06&0.64&0.40\\32&0.86&0.02&0.34&72&0.08&0.64&0.14\\33&0.90&0.00&0.28&73&0.06&0.82&0.36\\34&0.92&0.02&0.26&74&0.08&0.88&0.32\\35&0.92&0.04&0.32&75&0.02&0.86&0.34\\36&0.92&0.02&0.32&76&0.06&0.74&0.14\\37&0.90&0.00&0.40&77&0.02&0.92&0.42\\38&0.90&0.02&0.32&78&0.08&0.72&0.16\\39&0.82&0.02&0.38&79&0.04&0.76&0.42\\40&0.74&0.02&0.08&80&0.12&0.78&0.24\\
    \bottomrule
    
    \end{tabular}
    }
\end{table*}
\newpage
\subsection{Multi-Agent Debate Setting}

\paragraph{Communication Strategies(Proposed by ChatEval) and Topology Setting}
\begin{figure*}[!h]
  \includegraphics[width=1\linewidth]{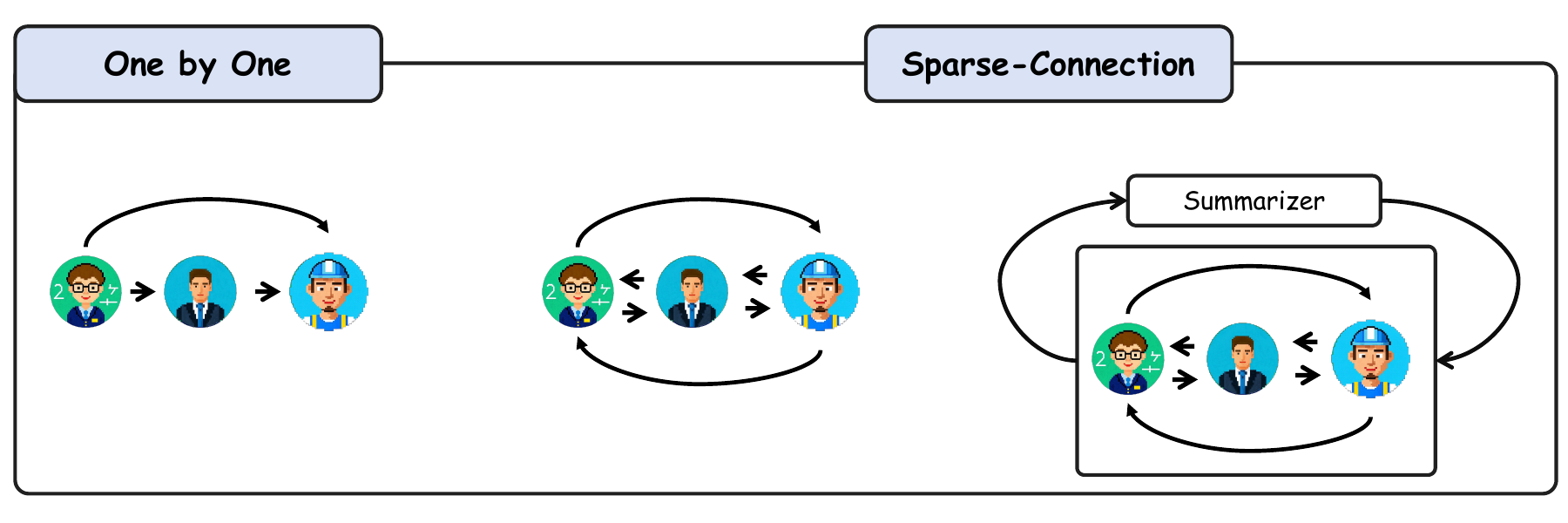} 
  
  \caption {Different communication strategies}
  \label{fig:pre_1}
\end{figure*}

1. One-By-One. During each round of the debate, the debater agents take turns in a set order to generate their response based on the current observation. When it’s time for a debater agent to respond, we directly concatenate what previous other agents have said into its chat history slot.

2. Simultaneous-Talk. Unlike the one-by-one strategy, we carry out an alternative communication strategy called simultaneous-talk, where debater agents are prompted to asynchronously generate responses in each iteration of the discussion to nullify the impact of the speaking order.

3. Simultaneous-Talk-with-Summarizer. The main difference between this strategy and simultaneous-talk is that we additionally employ another LLM as a summarizer. At the end of each iteration of the debate, we prompt this extra LLM to summarize the messages conveyed so far and concatenate this summarization into all debater agents’ chat history slots.

\begin{figure*}[!h]
  \includegraphics[width=1\linewidth]{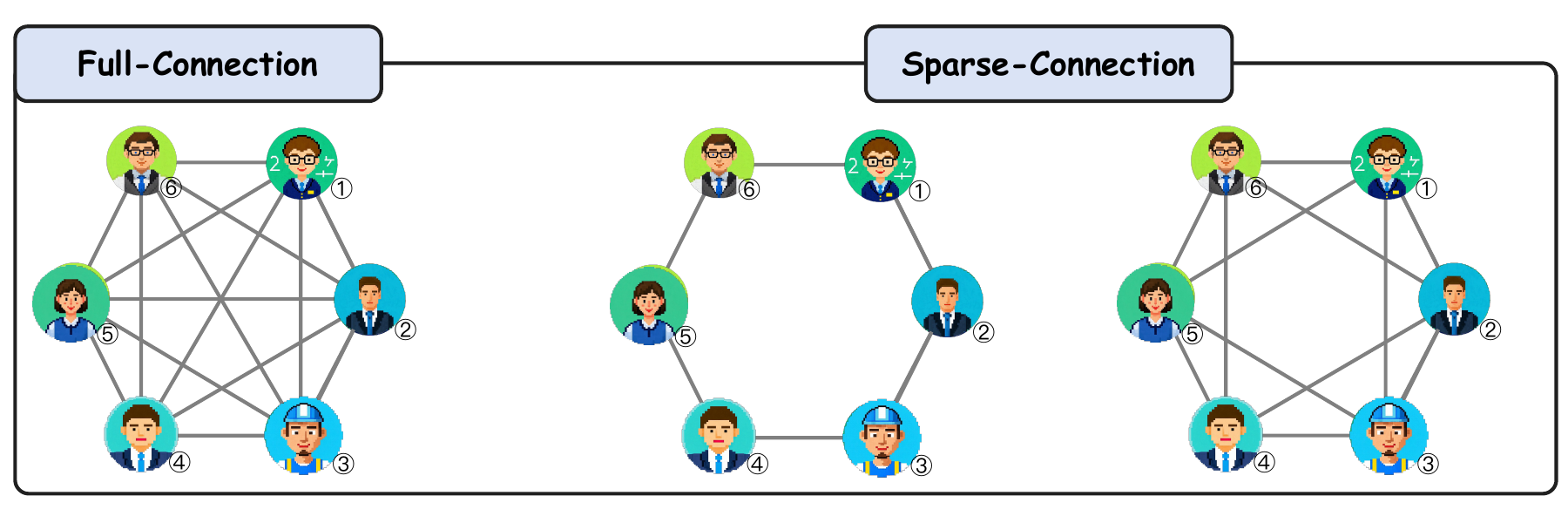} 
  
  \caption {Full connection and sparse connection}
  \label{fig:pre_1}
\end{figure*}

% \newpage
\subsection{Discussion of Societal Impact}\label{sec:social_impact}
Our research on Multi-Agent Debate (MAD) and the Multi-Agent Debate Consistency (MADC) strategy highlights important social implications, particularly concerning power concentration in decision-making processes.

\paragraph{Power Concentration}
If the "last-position" agent (e.g., the final decision-maker) holds excessive influence, several issues may arise:

\quad (1)Systemic Bias: The dominance of a few agents can suppress diverse viewpoints, leading to skewed outcomes.

\quad (2)Hidden Manipulation: Developers could artificially bias outcomes by assigning preferred agents to critical positions, undermining the integrity of the debate.

To mitigate these risks, we propose four key measures:

\quad (1)Multi-Agent Consensus (MADC): This strategy requires agreement among multiple independent agents. By ensuring reliability through diverse reasoning paths, we can reduce the likelihood of biased outcomes.

\quad (2)Dynamic Role Switching: Allowing agents to alternate roles (e.g., from critic to advisor) helps prevent monopolization and encourages a wider range of viewpoints, thereby enhancing the quality of the debate.

\quad (3)Power Decay: Agents allocated to key roles gradually lose weighting priority over time. This approach addresses the long-term concentration of decision-making power and promotes a more equitable distribution of influence.

\quad (4)Adversarial Training: To ensure model security and algorithm fairness, it is crucial to test positional sensitivity during training. This helps detect and reduce manipulation risks, ensuring that the system remains robust against biases.

\paragraph{Extreme Situations} In scenarios where Multi-Agent (MA) systems simulate human social behaviors, significant systemic biases can emerge, particularly in the "last-position" role. Many existing works on social simulation MA systems have not accounted for the role allocation factors discussed in this paper. For instance, in a simulation of the United Nations General Assembly, the speaking order of each country is often fixed. Similarly, in a simulated court, the plaintiff speaks first, followed by the defendant, with the judge summarizing last. In extreme cases, if there is only one round of debate, the judge's summary may consistently lean in favor of the defendant, leading to unfair outcomes.

By addressing these issues through our proposed measures, we can enhance the effectiveness and fairness of multi-agent systems in simulating social interactions and decision-making processes. This will contribute to more balanced and reliable outcomes in various applications, from governance to public discourse.
\end{document}